\documentclass[letterpaper]{article} 
\usepackage{aaai25}  
\usepackage{times}  
\usepackage{helvet}  
\usepackage{courier}  
\usepackage[hyphens]{url}  
\usepackage{graphicx} 
\usepackage{natbib}  
\usepackage{caption} 
\usepackage{algorithm}
\usepackage{algorithmic}
\usepackage{amsmath}
\usepackage{appendix}

\usepackage[linesnumbered,ruled,vlined,algo2e]{algorithm2e}

\usepackage{newfloat}
\usepackage{listings}
\DeclareCaptionStyle{ruled}{labelfont=normalfont,labelsep=colon,strut=off} 
\floatstyle{ruled}
\newfloat{listing}{tb}{lst}{}
\floatname{listing}{Listing}
\nocopyright

\title{Concept Conductor: Orchestrating Multiple Personalized Concepts in Text-to-Image Synthesis}
\author{
    Zebin Yao,
    Fangxiang Feng,
    Ruifan Li,
    Xiaojie Wang
}
\affiliations{
    Beijing University of Posts and Telecommunications\\

    \{zebin.yao, fxfeng, rfli, xjwang\}@bupt.edu.cn
}

\usepackage{bibentry}

\begin{document}

\maketitle

\begin{abstract}
The customization of text-to-image models has seen significant advancements, yet generating multiple personalized concepts remains a challenging task. Current methods struggle with attribute leakage and layout confusion when handling multiple concepts, leading to reduced concept fidelity and semantic consistency. In this work, we introduce a novel training-free framework, Concept Conductor, designed to ensure visual fidelity and correct layout in multi-concept customization. Concept Conductor isolates the sampling processes of multiple custom models to prevent attribute leakage between different concepts and corrects erroneous layouts through self-attention-based spatial guidance. Additionally, we present a concept injection technique that employs shape-aware masks to specify the generation area for each concept. This technique injects the structure and appearance of personalized concepts through feature fusion in the attention layers, ensuring harmony in the final image. Extensive qualitative and quantitative experiments demonstrate that Concept Conductor can consistently generate composite images with accurate layouts while preserving the visual details of each concept. Compared to existing baselines, Concept Conductor shows significant performance improvements. Our method supports the combination of any number of concepts and maintains high fidelity even when dealing with visually similar concepts. The code and models are available at \url{https://github.com/Nihukat/Concept-Conductor}.
\end{abstract}

%

\section{Introduction}

Text-to-image diffusion models\cite{nichol2021glide,saharia2022photorealistic,ramesh2022hierarchical,rombach2022high,podell2023sdxl} have achieved remarkable success in generating realistic high-resolution images. Building on this foundation, techniques for personalizing these models have also advanced. Various methods for single-concept customization\cite{dong2022dreamartist,ruiz2023dreambooth,gal2022image,voynov2023p+,alaluf2023neural} have been proposed, enabling the generation of images of the target concept in specified contexts based on user-provided visual conditions. These methods allow users to place real-world subjects into imagined scenes, greatly enriching the application scenarios of image generation.

\begin{figure}[t]
\centering
\includegraphics[width=1.0\columnwidth]{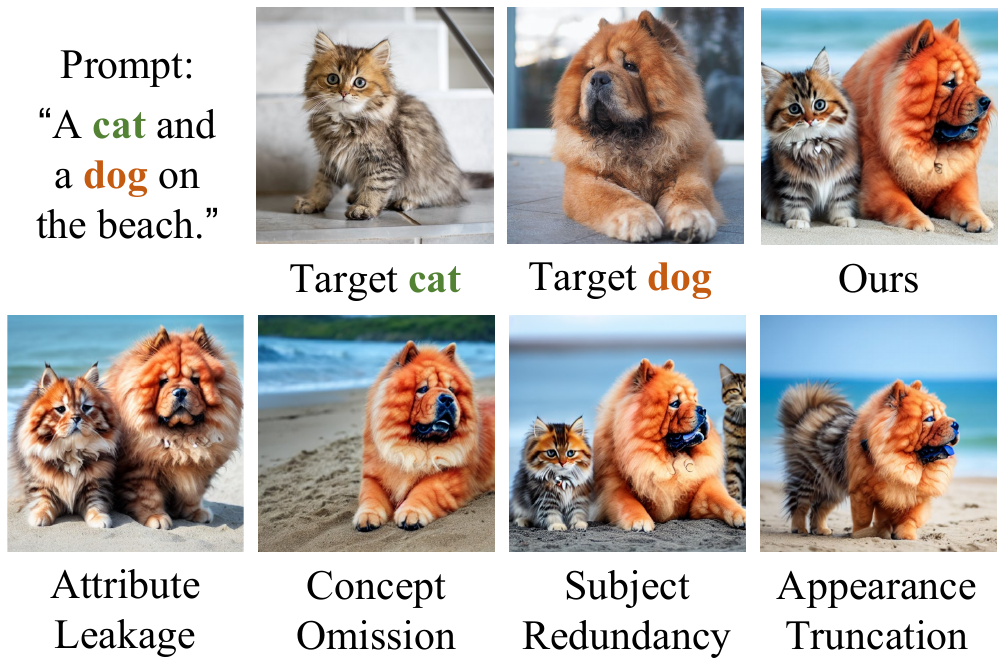} 
\caption{Results from existing multi-concept customization methods (second row) and our method (top right). Our method aims to address attribute leakage and layout confusion (concept omission, subject redundancy, appearance truncation), producing visually faithful and text-aligned images.}
\label{fig1}
\end{figure}

Despite the excellent performance of existing methods for single-concept customization, handling multiple concepts remains challenging. Current methods\cite{kumari2023multi,liu2023cones2,gu2024mix} often mix the attributes of multiple concepts or fail to align well with the given text prompts, especially when the target concepts are visually similar (e.g., a cat and a dog). We categorize these failures as attribute leakage and layout confusion. Layout confusion can be further divided into concept omission, subject redundancy, and appearance truncation, as shown in Figure \ref{fig1}. \textbf{Attribute leakage} denotes the application of one concept’s attributes to another (e.g., a cat acquiring the fur and eyes of a dog). \textbf{Concept omission} indicates one or more target concepts not appearing in the image (e.g., the absence of the target cat). \textbf{Subject redundancy} refers to the appearance of extra subjects similar to the target concept (e.g., an extra cat). \textbf{Appearance truncation} signifies the target concept’s appearance being observed only in a partial area of the subject (e.g., the upper half of a dog and the lower half of a cat).

To address these challenges, we introduce Concept Conductor, a novel inference framework for multi-concept customization. This framework aims to seamlessly integrate multiple personalized concepts with accurate attributes and layout into a single image based on the given text prompts, as illustrated in Figure \ref{fig2}. Our method comprises three key components: multipath sampling, layout alignment, and concept injection. \textbf{Multipath sampling} allows the base model and different single-concept models to retain their independent denoising processes, thereby preventing attribute leakage between concepts. Instead of training a model containing multiple concepts to directly generate the final image, we let each single-concept model first focus on generating its corresponding concept. These generated subjects are then integrated into a single image, avoiding interference and conflict between concepts. \textbf{Layout alignment} ensures each model produces the correct layout, fundamentally addressing layout confusion. Specifically, we borrow the layout from a normal image and align the intermediate representations produced by each model in the self-attention layers with it. This reference image is flexible and easy to obtain. It can be a real photo, generated by advanced text-to-image models, or even a simple collage created by the user. \textbf{Concept injection} enables each concept to fully inject its visual features into the final generated image, ensuring harmony. We use shape-aware masks to define the generation area for each concept and inject the visual details (including structure and appearance) of personalized concepts through feature fusion in the attention layers. At each step of the denoising process, we first use layout alignment to correct the input latent space representation, and then use concept injection to obtain the next representation. Multipath sampling is implemented in both layout alignment and concept injection to ensure the independence of each subject and coordination between different subjects.

To evaluate the effectiveness of the proposed method, we create a new dataset containing 30 concepts, covering representative categories such as humans, animals, objects, and buildings. We also introduce a fine-grained metric specifically designed for multi-concept customization to measure the visual consistency between the generated subjects and the given concepts. Extensive experiments demonstrate that our method can consistently generate composite images with correct layouts while fully preserving the attributes of each personalized concept, regardless of the number or similarity of the target concepts. Both qualitative and quantitative comparisons highlight the advantages of our method in terms of concept fidelity and alignment with textual semantics.

Our contributions can be summarized as follows:

\begin{itemize}
    \item We introduce Concept Conductor, a novel framework for multi-concept customization, preventing attribute leakage and layout confusion through multipath sampling and self-attention-based spatial guidance.
    \item We develop a concept injection technique, utilizing shape-aware masks and feature fusion to ensure harmony and visual fidelity in multi-concept image generation.
    \item We construct a new dataset containing 30 personalized concepts across representative categories. Comprehensive experiments on this dataset demonstrate the superior concept fidelity and alignment with textual semantics of the proposed Concept Conductor, including for concepts that are visually similar.
\end{itemize}

\section{Related Work}

\subsection{Text-to-Image Diffusion Models}

In recent years, text-to-image diffusion models have excelled in generating realistic and diverse images, becoming the mainstream approach in this field. Trained on large-scale datasets like LAION\cite{schuhmann2022laion}, models such as GLIDE\cite{nichol2021glide}, DALL-E 2\cite{ramesh2022hierarchical}, Imagen\cite{saharia2022photorealistic}, and Stable Diffusion\cite{rombach2022high} can produce high-quality and text-aligned outputs. However, these models struggle to understand the relationships between multiple concepts, resulting in generated content that fails to fully convey the original semantics. This issue is exacerbated when dealing with visually similar concepts. In this work, we apply our method to the publicly available Stable Diffusion\cite{rombach2022high}, which is based on the Latent Diffusion Model (LDM) architecture. LDM operates in the latent space of a Variational Autoencoder (VAE), iteratively denoising to recover the latent representation of an image from Gaussian noise. At each timestep, the noisy latents $z_{t}$ are fed into the denoising network $\epsilon_{\theta}$, which predicts the current noise $\epsilon_{\theta}(z_{t},y,t)$ based on the encoded prompt $y$.

\subsection{Customization in T2I Diffusion Models}

Several works\cite{ruiz2023dreambooth,gal2022image,voynov2023p+,alaluf2023neural} have customized text-to-image models to generate images of target concepts in new contexts by learning new visual concepts from user-provided example images. For instance, DreamBooth\cite{ruiz2023dreambooth} embeds specific visual concepts into a pre-trained model by fine-tuning its weights. Textual Inversion\cite{gal2022image} represents new concepts by optimizing a text embedding, later improved by P+\cite{voynov2023p+} and NeTI\cite{alaluf2023neural}. Custom Diffusion\cite{kumari2023multi} explores multi-concept customization through joint training, but it requires training a separate model for each combination and often faces severe attribute leakage. Recent works\cite{liu2023cones2,gu2024mix} propose frameworks that combine multiple single-concept models and introduce manually defined layouts in attention maps to aid generation. For example, Cones 2\cite{liu2023cones2} proposes residual embedding-based concept representations for textual composition and emphasizes or de-emphasizes a concept at a specific location by editing cross-attention. Mix-of-show\cite{gu2024mix} merges multiple custom models into one using gradient fusion and restricts each concept’s appearance area through region-controlled sampling. These works cannot fully avoid interference between concepts, leading to low success rates when handling similar concepts or more than two concepts. Additionally, by only manipulating cross-attention and neglecting the impact of self-attention on image structure, these methods often result in mismatched structures and appearances, causing layout control failures. In contrast, our method prevents attribute leakage by isolating the sampling processes of different single-concept models and achieves stable layout control through self-attention-based spatial guidance.

\subsection{Spatial Control in T2I Diffusion Models}

Using text prompts alone is insufficient for precise control over image layout or structure. Some methods\cite{avrahami2023spatext,li2023gligen,zhang2023adding,mou2024t2i} introduce layout conditions by training additional modules to generate controllable images. For example, GLIGEN\cite{li2023gligen} adds trainable gated self-attention layers to allow extra input conditions, such as bounding boxes. To achieve finer spatial control, ControlNet\cite{zhang2023adding} and T2I-Adapter\cite{mou2024t2i} introduce image-based conditions like keypoints, sketches, and depth maps by training U-Net encoder copies or adapters. These methods can stably control image structure but limit the target subjects’ poses, and these conditional images are difficult for users to create. Some training-free methods achieve spatial guidance by manipulating attention layers during sampling. Most works\cite{ma2024directed,kim2023dense,he2023localized} attempt to alleviate attribute leakage and control layout by directly editing attention maps but have low success rates. Several gradient-based methods \cite{couairon2023zero,xie2023boxdiff,phung2024grounded} calculate the loss between attention and the given layout and introduce layout information into the latent space representation by optimizing the loss gradients. These methods align the generated image layout with coarse visual conditions (e.g., bounding boxes and semantic segmentation maps), often resulting in high-frequency detail loss and image quality degradation, even with complex loss designs. In this work, we propose extracting layout information from an easily obtainable reference image as a supervisory signal, which not only stably controls the layout but also preserves the diversity of the generated subjects’ poses while avoiding image distortion.

\section{Method}

\subsection{Preliminary: Attention Layers in LDM}

LDM employs a U-Net as the denoising network, consisting of a series of convolutional layers and transformer blocks. In each block, intermediate features produced by the convolutional layers are passed to a self-attention layer followed by a cross-attention layer. Given an input feature $h_{\text{in}}$, the output feature in each attention layer is computed as $h_{\text{out}}=AV$, where $A=\text{softmax}(QK^{T})$. Here, $Q=f_{Q}(h_{\text{in}})$, $K=f_{K}(c)$, and $V=f_{V}(c)$ are obtained through learned projectors $f_{Q}$, $f_{K}$, and $f_{V}$, with $c=h_{\text{in}}$ for self-attention and $c=y$ for cross-attention. Self-attention enhances the quality of the generated image by capturing long-range dependencies in the image features, while cross-attention integrates textual information into the generation process, enabling the generated image to reflect the content of the text prompt. Furthermore, extensive researches \cite{liu2024towards,patashnik2023localizing,hertz2022prompt} have shown that self-attention controls the structure of the image (e.g., shapes, geometric relationships), whereas cross-attention controls the appearance of the image (e.g., colors, materials, textures).

\subsection{Preliminary: ED-LoRA}

ED-LoRA is a method for single-concept customization, primarily involving learnable hierarchical text embeddings and low-rank adaptation (LoRA) applied to pre-trained weights. To learn the representation of a concept within the pre-trained model’s domain, it creates layer-wise embeddings for the target concept’s token following P+\cite{voynov2023p+}. Additionally, to capture out-of-domain visual details, it fine-tunes the pre-trained text encoder and U-Net using LoRA\cite{hu2021lora}. In this paper, ED-LoRA is used as our single-concept model by default.

\subsection{Overview of Concept Conductor}

\begin{figure}[t]
\centering
\includegraphics[width=1.0\columnwidth]{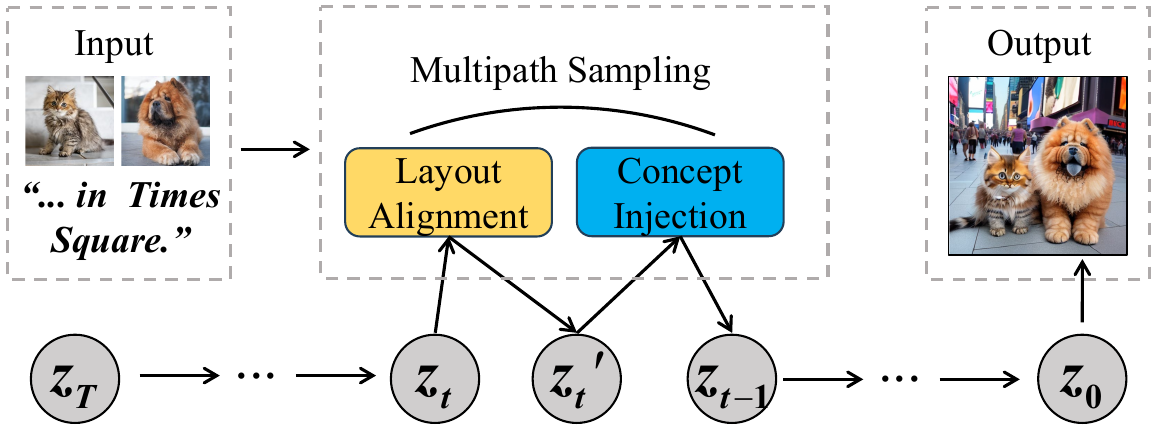} 
\caption{Overview of our proposed Concept Conductor. At each denoising step, the input latent vector $z_{t}$ is first corrected to $z_{t}\prime$ by the Layout Alignment module. $z_{t}\prime$ is then sent to the Concept Injection module for denoising, producing the next latent vector $z_{t-1}$. Both Layout Alignment and Concept Injection utilize the Multipath Sampling structure. After denoising, our method can generate images that align with the given text prompt and visual concepts.}
\label{fig2}
\end{figure}

\begin{figure}[t]
\centering
\includegraphics[width=1.0\columnwidth]{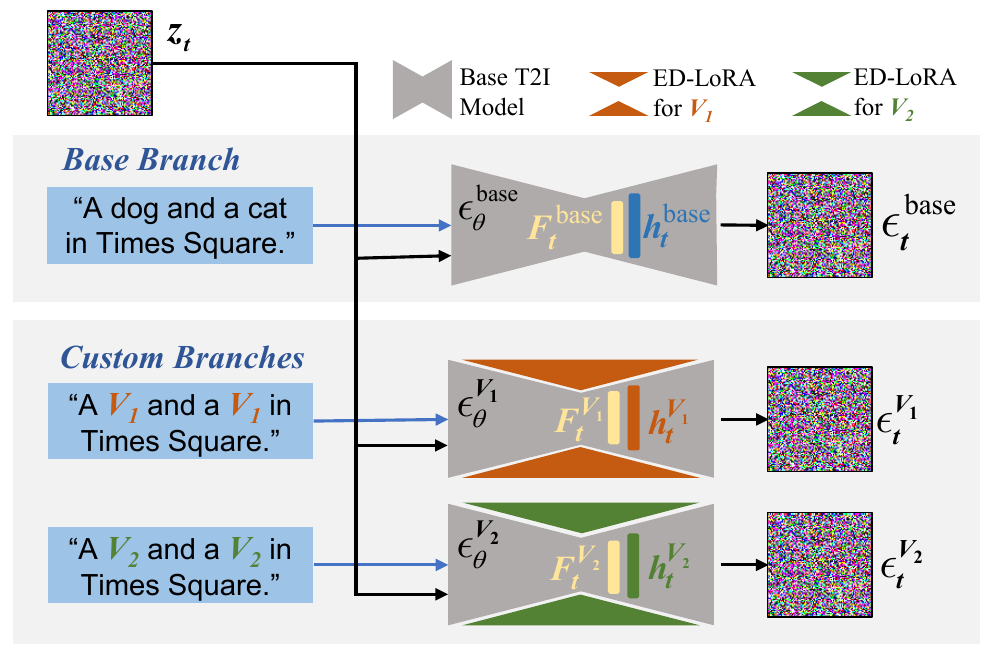} 
\caption{Illustration of multipath sampling. custom models $\epsilon_{\theta}^{V_{1}}$ and $\epsilon_{\theta}^{V_{2}}$ are created by adding ED-LoRA to the base model $\epsilon_{\theta}^{\text{base}}$. The base prompt and edited prompts are sent to the base model and custom models, respectively. Different models receive the same latent input $z_{t}$ and predict different noises. Self-attention features $F_{t}^{\text{base}}$, $F_{t}^{V_{1}}$, $F_{t}^{V_{2}}$, and the output feature maps of the attention layers $h_{t}^{\text{base}}$, $h_{t}^{V_{1}}$, $h_{t}^{V_{2}}$ are recorded during this process.
}
\label{fig3}
\end{figure}

\begin{figure}[t]
\centering
\includegraphics[width=1.0\columnwidth]{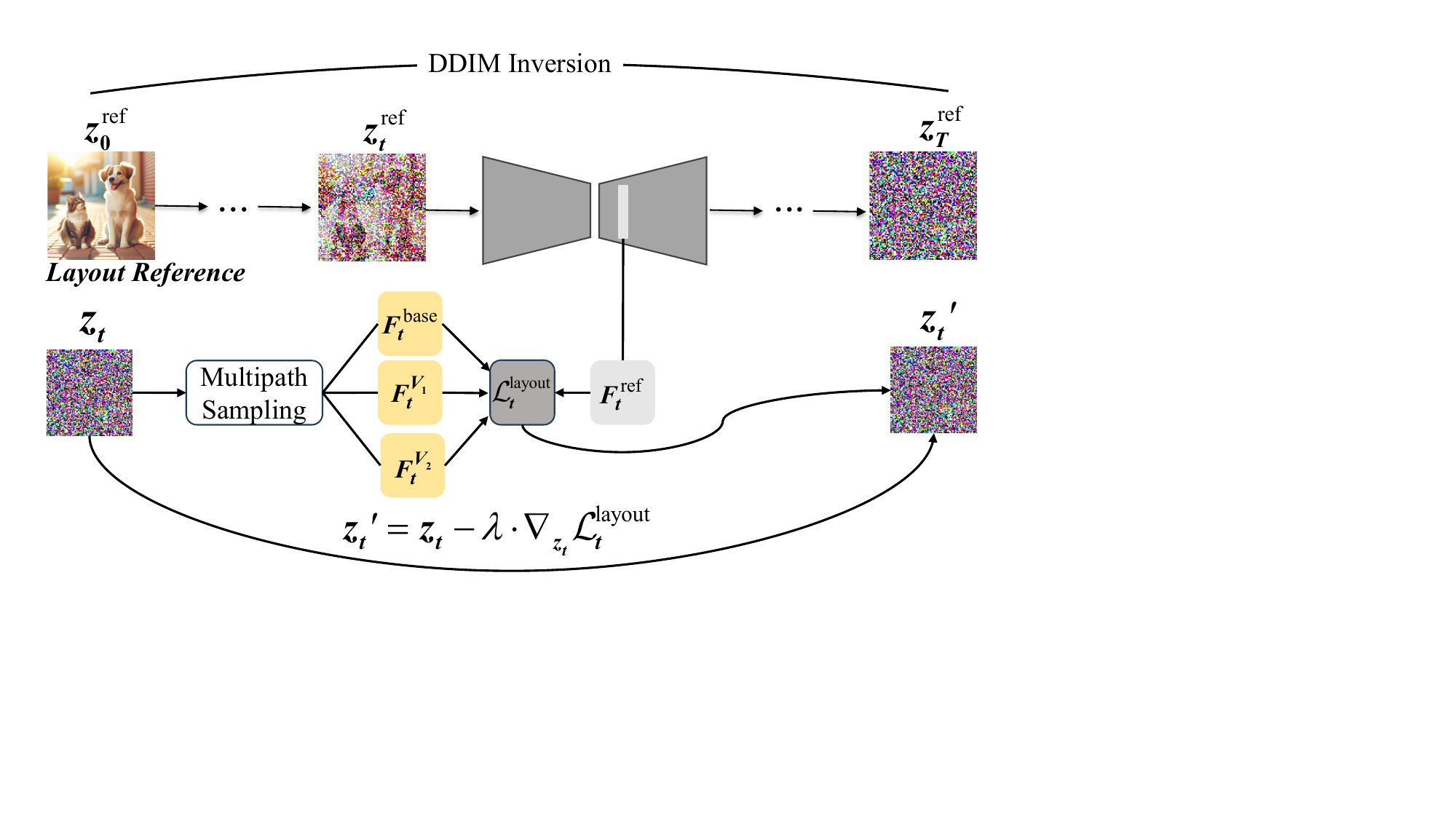} 
\caption{Illustration of layout alignment. The self-attention feature $F_{t}^{\text{ref}}$ of the layout reference image is extracted through DDIM inversion, which is then used to compute the loss with $F_{t}^{\text{base}}$, $F_{t}^{V_{1}}$, and $F_{t}^{V_{2}}$, updating the input latent vector $z_{t}$. For simplicity, the conversion from pixel space to latent space is omitted.
}
\label{fig4}
\end{figure}

\begin{figure}[t]
\centering
\includegraphics[width=1.0\columnwidth]{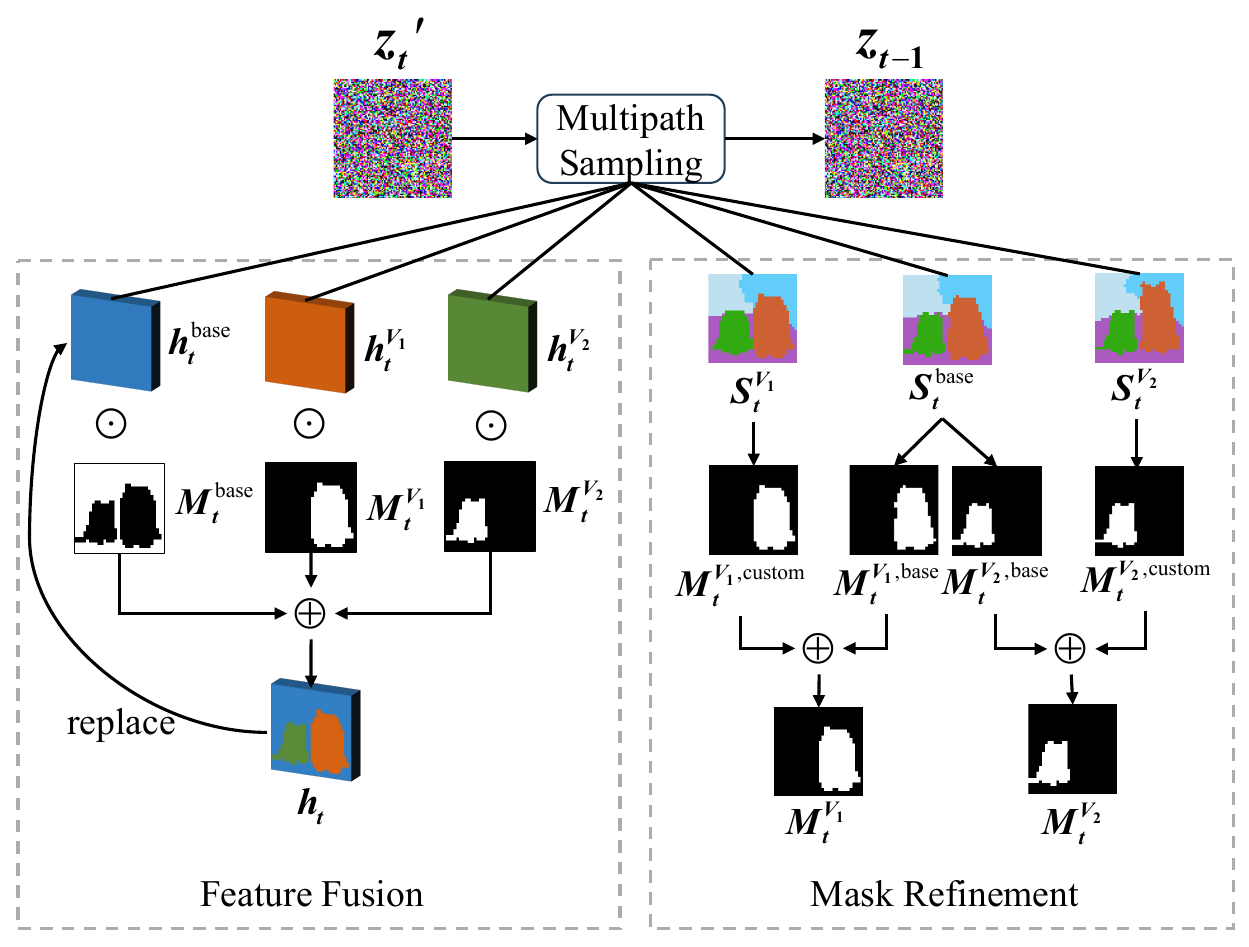} 
\caption{Illustration of concept injection, consisting of two parts: (1) Feature Fusion. The output feature maps of the attention layers from different models are multiplied by their corresponding masks and summed to obtain the fused feature map $h_{t}$, which is used to replace the original feature map $h_{t}^{\text{base}}$. (2) Mask Refinement. Segmentation maps are obtained by clustering on the self-attention, and the masks required for feature fusion are extracted from these maps.
}
\label{fig5}
\end{figure}

Our method comprises three components: multipath sampling, layout alignment, and concept injection, as illustrated in Figure \ref{fig2}. At each denoising step, we first correct the input latents $z_{t}$ through layout alignment, obtaining new latents $z_{t}\prime$ that carry the layout information from the reference image. Then, we inject the personalized concepts from the custom models into the base model and denoise $z_{t}\prime$ to obtain the next latents $z_{t-1}$. Multipath sampling is implemented in both layout alignment and concept injection to ensure the independence of each concept and coordination between different subjects.

\subsection{Multipath Sampling}

Joint training or model fusion methods often lead to attribute leakage between different concepts (as shown in Figure \ref{fig1}) and require additional optimization steps for each combination. To directly utilize multiple existing single-concept models for composite generation without attribute leakage, we propose a multipath sampling structure. This structure incorporates a base model $\epsilon_{\theta}^{\text{base}}$ and multiple custom models $\epsilon_{\theta}^{V_{i}}$ (implemented with ED-LoRA\cite{gu2024mix}) for personalized concepts $V_{i}$, as illustrated in Figure \ref{fig3}.

Given several custom models $\epsilon_{\theta}^{V_{i}}$ and a text prompt $p$, at each timestep $t$, we maintain the independent denoising process for each model: $\epsilon_{t}^{V_{i}}=\epsilon_{\theta}^{V_{i}}(z_{t}, t, p)$. When the prompt contains similar concepts, models may struggle to distinguish them, leading to attribute leakage. Therefore, we edit the input text prompt for each custom model to help them focus on generating the corresponding single concept. Given a base prompt $p_{\text{base}}$, we replace tokens visually similar to the target concept $V_{i}$ with tokens representing the target concept, creating a prompt variant $p_{V_{i}}$. For example, for the prompt “A dog and a cat on the beach” and concepts of a dog $V_{1}$ and a cat $V_{2}$, we edit the text to obtain two modified prompts: $p_{V_{1}}=$ “A $<V_{1}>$ and a $<V_{1}>$ on the beach” and $p_{V_{2}}=$ “A $<V_{2}>$ and a $<V_{2}>$ on the beach”.

After editing the prompts, the denoising process for the custom models can be expressed as: $\epsilon_{t}^{V_{i}}=\epsilon_{\theta}^{V_{i}}(z_{t},t,p_{V_{i}})$. Meanwhile, the base prompt is sent to the base model to retain global semantics: $\epsilon^{\text{base}}_{t}=\epsilon_{\theta}^{\text{base}}(z_{t},t,p_{\text{base}})$. Through multipath sampling, each custom model receives only information relevant to its corresponding concept, fundamentally preventing attribute leakage between different concepts.

\subsection{Layout Alignment}

Existing multi-concept customization methods often suffer from layout confusion (as shown in Figure \ref{fig1}), especially when the target concepts are visually similar or numerous. To address these challenges, we introduce a reference image to correct the layout during the generation process. For example, to generate an image of a specific dog and cat in a specific context, we only need a reference image containing an ordinary cat and dog. One simple approach to achieve layout control is to convert the reference image into abstract visual conditions (e.g., keypoints or sketches) and then use ControlNet\cite{zhang2023adding} or T2I-Adapter\cite{mou2024t2i} for spatial guidance, which limits variability and flexibility and may reduce the fidelity of the target concepts. Another approach is to directly inject the full self-attention of the reference image into the generation process, transferring the overall structure of the image\cite{hertz2022prompt,kwon2024concept}. This strictly limits the poses of the target subjects, reducing the diversity of the generated images. Additionally, it requires structural similarity between the reference image subjects and target concepts to avoid distortions caused by shape mismatches.

To align the layout while preserving the structure of the target concepts, we propose a gradient-guided approach, as shown in Figure \ref{fig4}. Given a reference image, we perform DDIM inversion to obtain the latent space representation $z_{t}^{\text{ref}}$ and record the self-attention features $F_{t}^{\text{ref}}$ at each timestep. Similarly, we record the self-attention features of the base model and each custom model during the generation process, denoted as $F_{t}^{\text{base}}$ and $F_{t}^{V_{i}}$, respectively, as shown in Figure \ref{fig2}. An optimization objective is set to encourage the generated image’s layout to align with the given layout:
\begin{equation}
\mathcal{L}_{t}^{\text{layout}}=\|F^{\text{base}}_{t}-F^{\text{ref}}_{t}\|_{2}+\alpha\frac{1}{N}\sum\limits_{i=1}^{N}{\|F_{t}^{V_{i}}-F_{t}^{\text{ref}}\|_{2}}
\end{equation}
where $\alpha$ represents the weighting coefficient, and $N$ denotes the number of personalized concepts. We use gradient descent to optimize this objective, obtaining the corrected latent space representation:
\begin{equation}
z_t\prime=z_t-\lambda \cdot \nabla_{z_t} \mathcal{L}_{t}^{\text{layout}}
\end{equation}
where $\lambda$ represents the gradient descent step size. Through layout alignment, we ensure that the generated image mimics the reference image’s layout without any confusion.

\subsection{Concept Injection}

After layout alignment, the original latents $z_{t}$ are replaced with the corrected latents $z_{t}\prime$, and multipath sampling is used to generate the next latents $z_{t-1}$. The goal is to inject the subjects generated by different custom branches into the base branch to create a composite image. A naive way is to spatially fuse the noise predicted by different models: 
\begin{equation}
\epsilon_{t}^{\text{fuse}}=\epsilon_{t}^{\text{base}}\odot M^{\text{base}}+\sum\limits_{i=1}^{N}\epsilon_{t}^{V_{i}}\odot M^{V_{i}}
\end{equation}
where $\epsilon_{t}^{\text{base}}$ represents the noise predicted by the base model, $\epsilon_{t}^{V_{i}}$ represents the noise predicted by the custom model for concept $V_{i}$, and $M_{\text{base}}$ and $M_{V_{i}}$ are predefined masks. This method ensures the fidelity of the target concepts but often results in disharmonious images.

To address this issue,  we propose an attention-based concept injection technique, including feature fusion and mask refinement, as shown in Figure \ref{fig5}. Spatial fusion is implemented on the output feature maps of all attention layers in the U-Net decoder, as self-attention controls the structure of the subjects and cross-attention controls their appearance, both crucial for reproducing the attributes of the target concepts. For each selected attention layer, the fused output feature is computed as: 
\begin{equation}
h_{t}=h_{t}^{\text{base}}\odot M_{t}^{\text{base}}+\sum\limits_{i=1}^{N}h_{t}^{V_{i}}\odot M_{t}^{V_{i}}
\end{equation}
where $M_{t}^{\text{base}}=1-\bigcup\limits_{i=1}^{N}M_{t}^{V_{i}}$. Here, $h_{t}^{\text{base}}$ and $h_{t}^{V_{i}}$ represent the output features of the attention layers of the base model and the custom models, respectively, and $M_{t}^{V_{i}}$ represents the binary mask of concept $V_{i}$ at timestep $t$, specifying the dense generation area of the target concept. The fused feature $h_{t}$ is then sent back to the corresponding position in the base model to replace $h_{t}^{\text{base}}$ and complete the denoising process.

Since the poses of the generated subjects are uncertain, predefined masks may not precisely match the shapes and contours of the target subjects, leading to incomplete appearances. To address this, we use mask refinement to allow the masks to adjust according to the shapes and poses of the target subjects during the generation process. Inspired by local-prompt-mixing\cite{patashnik2023localizing}, we use self-attention-based semantic segmentation to obtain the masks of the target subjects. For each target concept $V_{i}$, we cluster the self-attention $A_{t}^{V_{i}}$ of the custom model $\epsilon_{V_{i}}$ to obtain a semantic segmentation map $S_{t}^{V_{i}}$ and extract the subject’s mask $M_{t}^{V_{i},\text{custom}}$. We perform the same operation on the self-attention $A_{t}^{\text{base}}$ of the base model $\epsilon_{\text{base}}$, obtaining the semantic segmentation map $S_{t}^{\text{base}}$ and several masks $M_{t}^{V_{i},\text{base}}$, $i\in[1,N]$, each corresponding to a subject $V_{i}$. To reconcile the shape differences between the subjects in the base model and the custom models, the corresponding masks are merged: $M_{t}^{V_{i}}=M_{t}^{V_{i}, \text{custom}}\cup M_{t}^{V_{i},\text{base}}$.

For initialization, we perform DDIM inversion on the reference image and extract the original masks $M_{T}^{V_{i}}$ from the self-attention layers in the same way. An alternative way is to use grounding models(e.g., Grounding DINO\cite{liu2023grounding}) and segmentation models(e.g., SAM\cite{kirillov2023segment}) to extract masks in the pixel space, which provides higher resolution masks but requires additional computation and storage overhead. Through concept injection, we ensure the harmony of the image while fully preserving the attributes of the target concepts.

\section{Experiments}

\begin{figure*}[t]
\centering
\includegraphics[width=1.5\columnwidth]{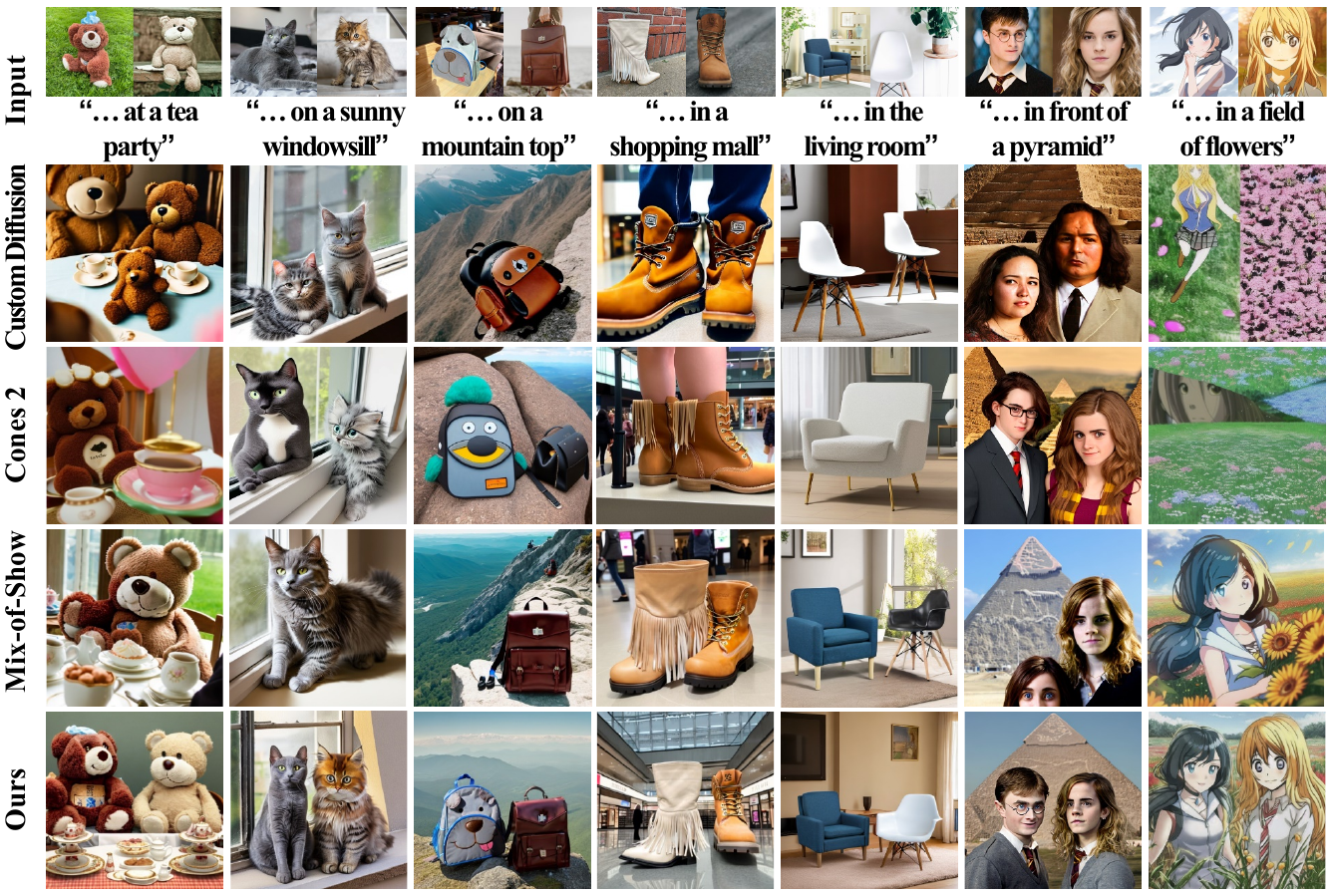} 
\caption{Qualitative comparison of multi-concept customization methods. The results show that our method ensures visual fidelity and correct layout, while other methods suffer from severe attribute confusion and layout disorder.
}
\label{fig6}
\end{figure*}

\subsection{Dataset}

We construct a dataset covering representative categories such as humans, animals, objects, and buildings, including 30 personalized concepts. Real and anime human images are collected from the Mix-of-show dataset\cite{gu2024mix}, while other categories are sourced from the DreamBooth dataset\cite{ruiz2023dreambooth} and CustomConcept101\cite{kumari2023multi}, with 3-15 images per concept. For quantitative evaluation, we select 10 pairs of visually similar concepts and generate 5 text prompts for each pair using ChatGPT\cite{chatgpt}. We produce 8 samples for each text prompt using the same set of random seeds, resulting in a total of 10×5×8=400 images per method. More details about the dataset are provided in Appendix A.1.

\subsection{Implementation Details}

Our method is implemented on Stable Diffusion v1.5, using images generated by SDXL\cite{podell2023sdxl} as layout references. In layout alignment, the key of the first self-attention layer in the U-Net decoder is used as the layout feature $F_{t}$. For mask refinement, we cluster the attention probabilities in the sixth self-attention layer of the U-Net decoder to extract semantic segmentation maps and scale them to different sizes for feature fusion in different attention layers. In all experiments, the weighting coefficient $\alpha$ is set to 1, and the gradient descent step size $\lambda$ is set to 10. More implementation details are provided in Appendix A.2.

\subsection{Baselines}

We compare our method with three multi-concept customization methods: Custom Diffusion, Cones 2, and Mix-of-Show. For Custom Diffusion, we use the diffusers\cite{von-platen-etal-2022-diffusers} version implementation, while for the other methods, we use their official code implementations. All experimental settings follow the official recommendations. For Cones 2 and Mix-of-Show, grounding models are used to extract bounding boxes of target subjects from the layout reference image to ensure consistent spatial conditions. To ensure a fair comparison, no additional control models like ControlNet or T2I-Adapter are used.

\subsection{Evaluation Metrics}

We evaluate multi-concept customization methods from two perspectives: text alignment and image alignment. For text alignment, we report results on CLIP\cite{radford2021learning} and ImageReward\cite{xu2024imagereward}. For image alignment, we introduce a new metric called Segmentation Similarity (SegSim) to address the limitations of traditional image similarity methods, which cannot reflect attribute leakage and layout conflicts. SegSim evaluates fine-grained fidelity by using text-guided grounding models and segmentation models to extract subject segments from generated and reference images, then calculating their similarity. Detailed information is in Appendix A.3. We use CLIP\cite{radford2021learning} and DINOv2\cite{oquab2023dinov2} to calculate segment similarity and report image alignment based on these models. To systematically evaluate omission and redundancy in multi-concept generation, grounding models are used to automatically count the number of target category subjects in each generated image.

\subsection{Qualitative Comparison}

\begin{figure}[t]
\centering
\includegraphics[width=1.0\columnwidth]{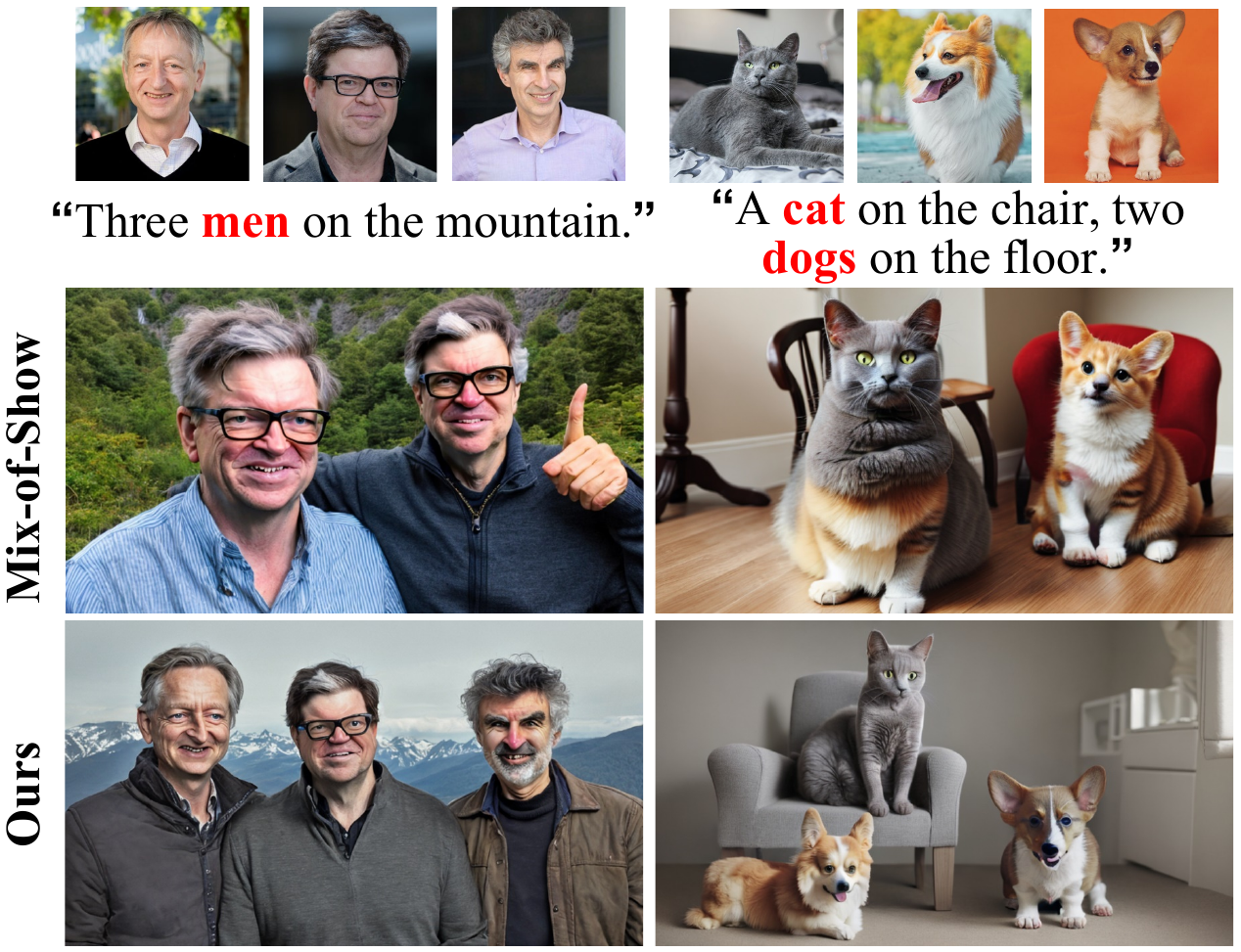} 
\caption{Qualitative comparison in challenging scenarios. Mix-of-Show struggles to handle more than two similar concepts or complex layouts, whereas our method demonstrates robust performance even in these complex scenarios.
}
\label{fig7}
\end{figure}

We evaluate our method and all baselines on various combinations of similar concepts, as shown in Figure \ref{fig6}. Custom Diffusion and Cones 2 struggle to retain the visual details of target concepts (e.g., the cat’s fur pattern and the backpack’s design), exhibiting severe attribute leakage (e.g., two identical boots) and layout confusion (e.g., missing or redundant teddy bears). Mix-of-Show demonstrates higher fidelity and mitigates attribute leakage but still faces significant concept omission (e.g., missing cartoon backpack) and appearance truncation (e.g., stitched teddy bear). In contrast, our Concept Conductor generates all target concepts with high fidelity without leakage through multipath sampling and concept injection, ensuring correct layout through layout alignment. Our method maintains stable performance across different concept combinations, even when the target concepts are very similar, such as two teddy bears.

We further explore more challenging scenarios and compare our method with Mix-of-Show, as shown in Figure \ref{fig7}. When handling three similar concepts, Mix-of-Show exhibites severe attribute leakage (e.g., both men wearing glasses) and concept omission (e.g., one person missing). Additionally, Mix-of-Show struggles with dense layouts, often resulting in appearance truncation when faced with complex spatial relationships (e.g., the upper half of a cat and the lower half of a dog stitched together). In contrast, our method maintains the visual features of each concept without attribute leakage and faithfully reflects the layout described in the text, even in these complex scenarios. 

\subsection{Quantitative Comparison}

\begin{table}[t]
\small
\setlength{\tabcolsep}{0.7mm}
\begin{tabular}{lcclcclcc}
\hline & \multicolumn{2}{c}{ Text-alignment } && \multicolumn{2}{c}{ Image-alignment }  && \multicolumn{2}{c}{ Counting }\\
\cline { 2 - 3 } \cline { 5 - 6 } \cline{ 8 - 9 } Method & CLIP-T  & IR && CLIP-I & DINO && $n<2$ &  $n>2$  \\
\hline CD & 0.2939 & 0.0133 && 0.8333 & 0.6829 && 0.2750 & 0.1050 \\
Cones2 & 0.2954 & 0.0973 && 0.8415 & 0.6928 && 0.3600 & 0.0650 \\
MoS & 0.2962 & 0.3944 && 0.8928 & 0.7961 && 0.3400 & 0.0400 \\
\hline Ours & $\mathbf{0.3107}$ & $\mathbf{1.2542}$ && $\mathbf{0.9190}$ & $\mathbf{0.8569}$  && $\mathbf{0.0075}$ & $\mathbf{0.0350}$ \\
\hline
\end{tabular}
\caption{Quantitative Comparison of Multi-Concept Customization Methods. IR stands for ImageReward, CD for Custom Diffusion, and MoS for Mix-of-Show. $n<2$ indicates omission, while $n>2$ indicates redundancy.}
\label{table1}
\end{table}

As reported in Table \ref{table1}, our Concept Conductor significantly outperforms previous methods in both image alignment and text alignment. The improvement in image alignment indicates that our method can preserve the visual details of each concept without attribute leakage, primarily due to our proposed multipath sampling framework and attention-based concept injection. The improvement in text alignment is mainly because our method effectively avoids the layout confusion that leads to unfaithful or disharmonious images through layout alignment, thereby enhancing text-image consistency. The significant reduction in omission and redundancy rates also supports this.

\subsection{Ablation Study}

\begin{figure}[t]
\centering
\includegraphics[width=1.0\columnwidth]{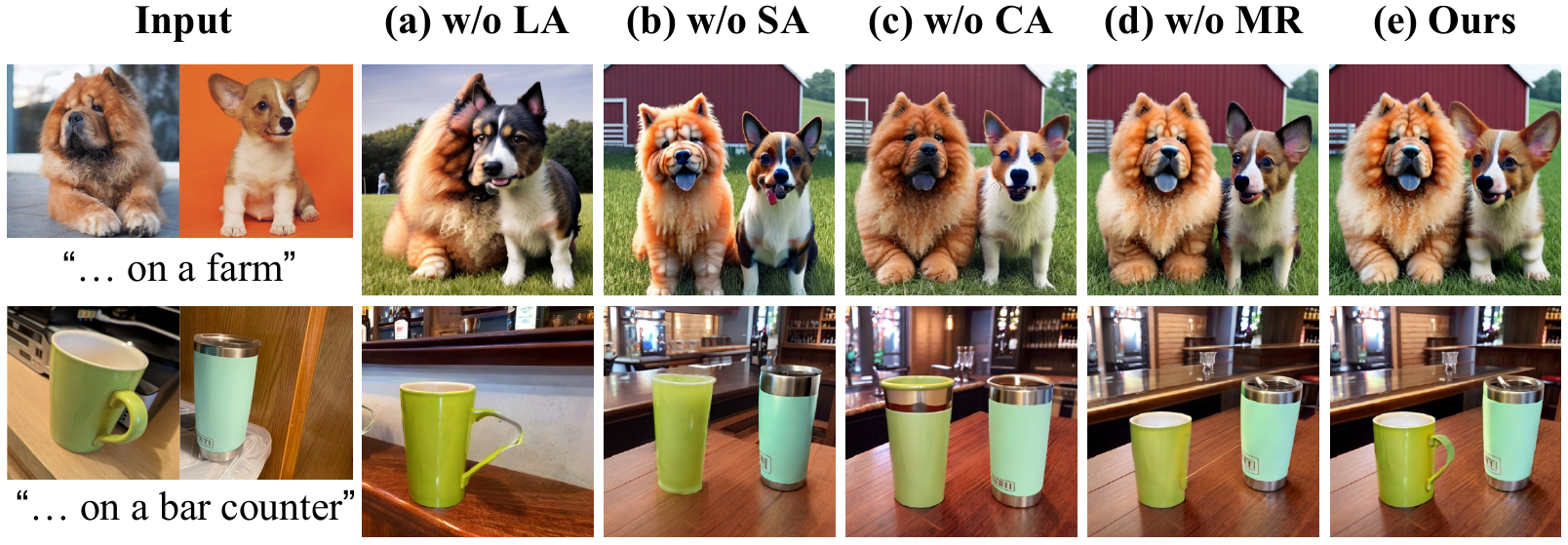} 
\caption{Qualitative comparison of ablation variants. (a) Results without layout alignment (LA). (b) Results without self-attention (SA) features in concept injection. (c) Results without cross-attention (CA) features in concept injection. (d) Results without mask refinement (MR) in concept injection. (e) Results of our complete model.
}
\label{fig8}
\end{figure}

\begin{table}[t]
\small
\setlength{\tabcolsep}{0.68mm}
\begin{tabular}{lcclcclcc}
\hline & \multicolumn{2}{c}{ Text-alignment } && \multicolumn{2}{c}{ Image-alignment }  && \multicolumn{2}{c}{ Counting }\\
\cline { 2 - 3 } \cline { 5 - 6 } \cline{ 8 - 9 } Method & CLIP-T  & IR && CLIP-I & DINO && $n<2$ &  $n>2$  \\
\hline w/o LA & 0.3020 & 0.8926 && 0.9021 & 0.8258 && 0.0575 & 0.0625 \\
w/o SA & 0.3095 & 1.2224 && 0.8940 & 0.7820 && 0.0450 & 0.0475 \\
w/o CA & 0.3027 & 1.2059 && 0.9051 & 0.8194 && 0.0275 & 0.0550 \\
w/o MR & 0.3034 & 1.2140 && 0.9148 & 0.8478 && 0.0100 & 0.0375 \\
\hline Ours & $\mathbf{0.3107}$ & $\mathbf{1.2542}$ && $\mathbf{0.9190}$ & $\mathbf{0.8569}$  && $\mathbf{0.0075}$ & $\mathbf{0.0350}$ \\
\hline
\end{tabular}
\caption{Quantitative Comparison of Ablation Variants. Ablation is performed on four components: Layout Alignment (LA), Self-Attention (SA), Cross-Attention (CA), and Mask Refinement (MR).}
\label{table2}
\end{table}

To verify the effectiveness of the proposed components, we conduct qualitative comparisons of various settings, as shown in Figure \ref{fig8}. In Figure \ref{fig8}(a), removing layout alignment leads to incorrect layouts, including appearance truncation (e.g., two dogs incorrectly stitched together) and concept omission (e.g., missing turquoise cup). Figures 8(b) and 8(c) show that disabling either self-attention or cross-attention features during concept injection results in a decline in fidelity, indicating that both are crucial for preserving the visual details of the target concepts. Figure \ref{fig8}(d) demonstrates that ablating mask refinement can lead to a mismatch between the generated subject contours and the target concepts, resulting in incomplete appearances (e.g., chow chow’s fur, corgi’s ears, green cup’s handle). To avoid randomness, we conduct quantitative comparisons of various settings using the same data and evaluation metrics as in the previous section, with results reported in Table \ref{table2}. As shown in Table \ref{table2}, layout alignment effectively avoids omission and redundancy, significantly improving the alignment of generated images with textual semantics. Feature fusion in both self-attention and cross-attention layers leads to higher image alignment, as both are crucial for reproducing the attributes of the target concepts. Mask refinement further improves text alignment and image alignment by optimizing the edge details of the generated subjects.

\section{Conclusion}

We introduce Concept Conductor, a novel inference framework designed to generate realistic images containing multiple personalized concepts. By employing multipath sampling and layout alignment, we addressed the common issues of attribute leakage and layout confusion in multi-concept personalization. Additionally, concept injection is used to create harmonious composite images. Experimental results demonstrate that Concept Conductor can consistently generate composite images with correct layouts, fully preserving the attributes of each concept, even when the target concepts are highly similar or numerous.

\bibliography{aaai25}

\clearpage
\section*{Appendix}

In this supplementary material, we provide additional details on our experimental procedures and analyses. In Appendix A, we describe the experimental settings in detail, including datasets, implementation details, and our proposed evaluation metric SegSim. In Appendix B, we present additional experimental results. In Appendix C, we analyze the limitations of our method. Finally, in Appendix D, we discuss the potential societal impacts of our approach.

\appendix

\section{A\quad Experimental Settings}

\subsection{A.1\quad Datasets}

\begin{figure}[h]
\centering
\includegraphics[width=1.0\columnwidth]{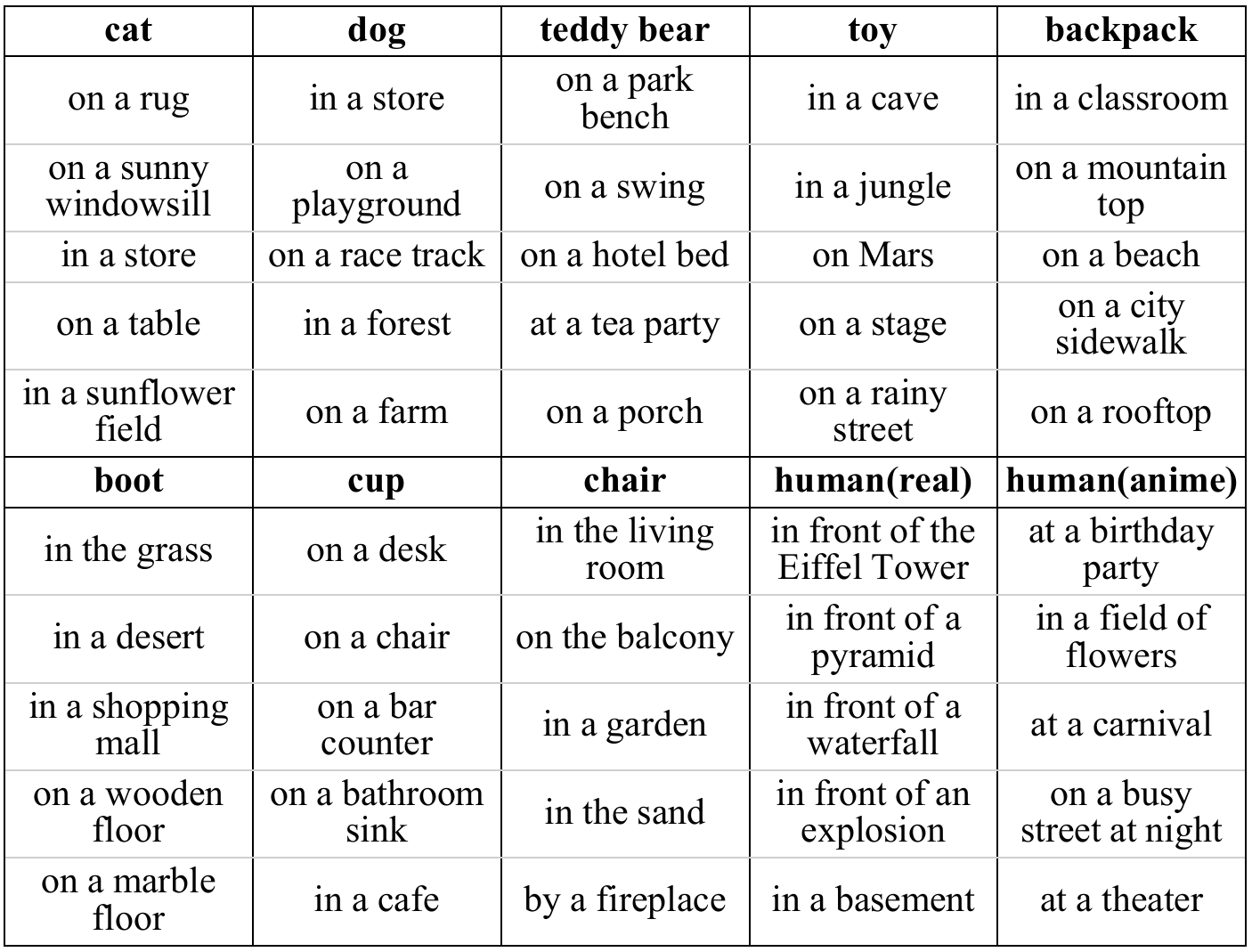} 
\caption{Scenes used in the prompts in quantitative evaluation, covering both indoor and outdoor settings. 
}
\label{fig9}
\end{figure}

\begin{figure}
\centering
\includegraphics[width=1.0\columnwidth]{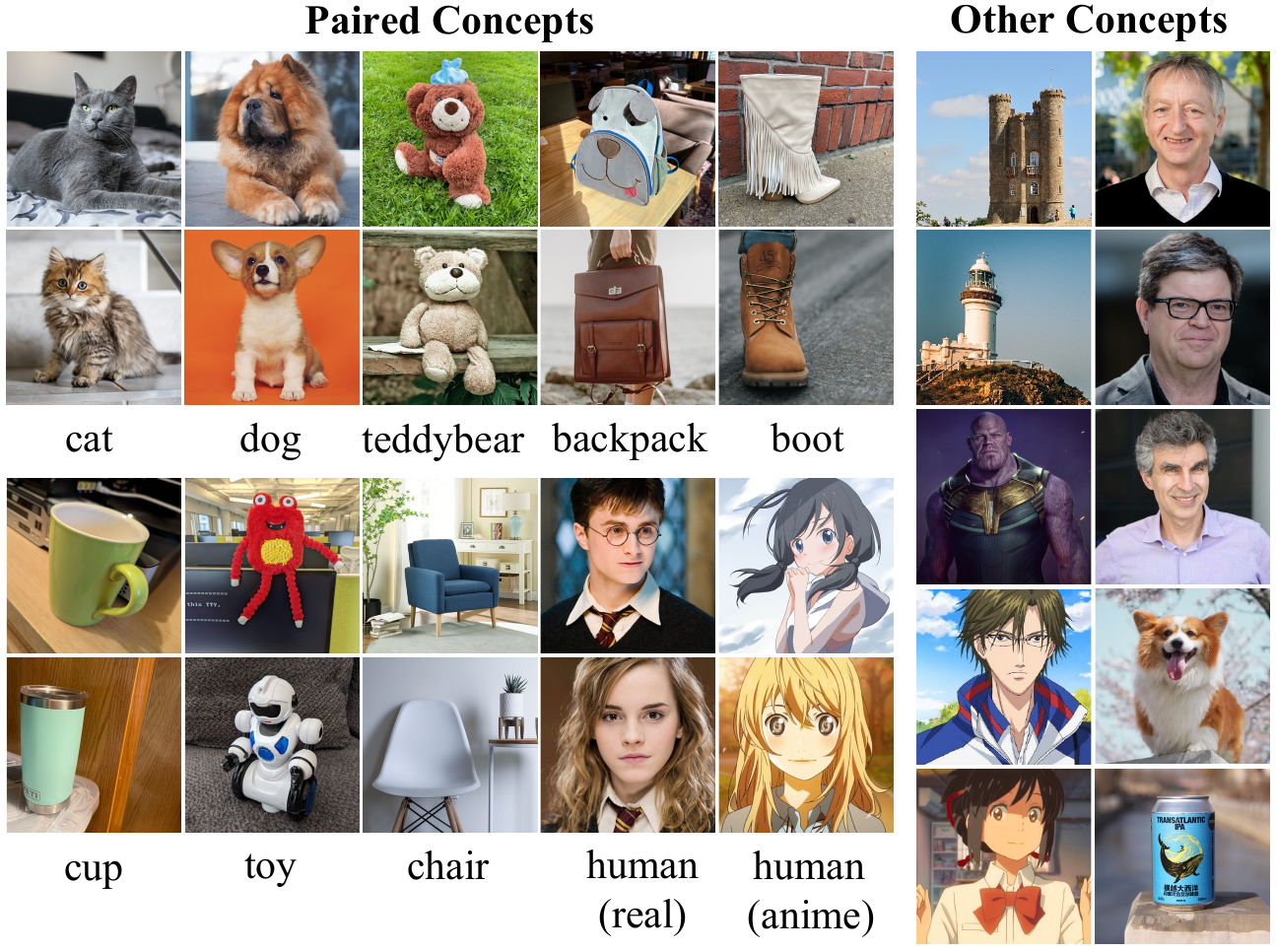} 
\caption{All personalized concepts used in this work. The left side shows paired concepts used in quantitative comparisons, while the right side shows concepts used in other experiments. 
}
\label{fig10}
\end{figure}

We select 30 personalized concepts from previous works\cite{ruiz2023dreambooth,kumari2023multi,gu2024mix}, including 6 real humans, 4 anime humans, 5 animals, 2 buildings, and 13 common objects. For quantitative evaluation, we choose 10 pairs of visually similar concepts, as summarized in Figure \ref{fig10}. We use ChatGPT\cite{chatgpt} to generate 5 text prompts for each pair of concepts. Each prompt includes two subjects and a scene (e.g., “Two toys on a stage.”). The scenes vary across different combinations, covering both indoor and outdoor settings, as detailed in Figure \ref{fig9}.

\begin{algorithm*}
\caption{Mask Refinement using Self-Attention Maps}
\KwIn{Personalized concepts $V = \{V_1, V_2, \ldots, V_n\}$, self-attention maps $A_{t}^{\text{base}}$ and $A_{t}^{V} = \{A_{t}^{V_1}, A_{t}^{V_2}, \ldots, A_{t}^{V_n}\}$, \\old masks $M_{t+1}^{V} = \{M_{t+1}^{V_{1}}, M_{t+1}^{V_{2}}, \ldots, M_{t+1}^{V_{n}}\}$ at timestep $t+1$}
\KwOut{Updated masks $M_{t}^{V} = \{M_{t}^{V_{1}}, M_{t}^{V_{2}}, \ldots, M_{t}^{V_{n}}\}$ at timestep $t$}

\For{each concept $V_{i} \in V$}{
    Apply clustering on $A_{t}^{V_{i}}$ using K-Means with cluster numbers from $|V|$ to $2|V|$ ;
    
    Record all segmentations $S_{t}^{V_{i},k}$ from each clustering $k$ ;
    
    $D(S_{t}^{V_{i},k}, M_{t+1}^{V_{i}}) = |S_{t}^{V_{i},k} \cap M_{t+1}^{V_{i}}| / |S_{t}^{V_{i},k} \cup M_{t+1}^{V_{i}}|$\tcp*{Compute matching degree }
    $k_{\text{max}} = \arg\max_{k} D(S_{t}^{V_{i},k}, M_{t+1}^{V_{i}})$\tcp*{Select the best matching segmentation}
    $M_{t}^{V_{i},\text{custom}}=S_{t}^{V_{i}, k_{\text{max}}}$ \;
    
    Apply clustering on $A_{t}^{\text{base}}$ using K-Means with cluster numbers from $|V|$ to $2|V|$;
    
    Record all segmentations $S_{t}^{\text{base},k}$ from each clustering $k$;
    
    $D(S_{t}^{\text{base},k}, M_{t+1}^{V_{i}}) = |S_{t}^{\text{base},k} \cap M_{t+1}^{V_{i}}| /  |S_{t}^{\text{base},k} \cup M_{t+1}^{V_{i}}|$\tcp*{Compute matching degree}
    $k_{\text{max}} = \arg\max_{k} D(S_{t}^{\text{base},k}, M_{t+1}^{V_{i}})$\tcp*{Select the best matching segmentation}
    $M_{t}^{V_{i},\text{base}} = S_{t}^{\text{base},k_{\text{max}}}$ ;
    
    $M_{t}^{V_{i}\prime} = M_{t}^{V_{i},\text{base}} \cup M_{t}^{V_{i},\text{custom}}$\tcp*{Combine base and custom model masks}
}
$M_{t}^{\text{sum}} = \sum_{i=1}^{N} M_{t}^{V_{i}\prime}$\tcp*{Sum all masks}
$\Omega_{t} = \begin{cases} 
1 & \text{if } M_{t}^{\text{sum}} > 1 \\
0 & \text{otherwise}
\end{cases}$\; \tcp*{Binarize the result to get overlapping regions}

\For{each concept $V_{i} \in V$}{
    $M_{t}^{V_{i}}=\Omega_{t}\odot M_{t+1}^{V_{i}}+(1-\Omega_{t})\odot M_{t}^{V_{i}\prime}$\tcp*{Replace overlapping regions with old masks}
}
\Return $M_{t}^{V} = \{M_{t}^{V_{1}}, M_{t}^{V_{2}}, \ldots, M_{t}^{V_{n}}\}$

\end{algorithm*}

\begin{algorithm*}
\caption{Segmentation Similarity (SegSim)}
\KwIn{Generated image $G$, Reference concepts $C = \{C_1, C_2, \ldots, C_n\}$, Prompts $P = \{p_1, p_2, \ldots, p_m\}$}
\KwOut{Image-alignment $Score$}
$G_{\text{segments}} = \{\}$ \tcp*{Initialize an empty set for generated segments}
\For{each prompt $p_i \in P$}{
    $segments = \text{extract\_segments}(G, p_i)$ \tcp*{Extract segments from $G$ using $p_i$}
    $G_{\text{segments}} = G_{\text{segments}} \cup segments$ \tcp*{Union of segments}
}
$concept\_similarities = []$ \tcp*{Initialize an empty list for concept similarities}
\For{each reference concept $C_i \in C$}{
    $group\_similarities = []$ \tcp*{Initialize an empty list for group similarities}
    \For{each reference image $R_{ij} \in C_i$}{
        $R_{ij\_\text{segments}} = \text{extract\_segments}(R_{ij}, p_i)$ \tcp*{Extract subject segments from $R_{ij}$ using a prompt}
        $max\_similarity = 0$ \tcp*{Initialize maximum similarity}
        \For{each segment $r \in R_{ij\_\text{segments}}$}{
            \For{each segment $g \in G_{\text{segments}}$}{
                 $sim = \text{calculate\_similarity}(r, g)$ \tcp*{Calculate similarity with pretrained models}
                \If{$sim > max\_similarity$}{
                    $max\_similarity = sim$ \tcp*{Update maximum similarity}
                }
            }
        }
        Append $max\_similarity$ to $group\_similarities$ \tcp*{Store maximum similarity}
    }
    $group\_average = \frac{1}{|C_i|} \sum_{j=1}^{|C_i|} group\_similarities[j]$ \tcp*{Calculate the average similarity}
    Append $group\_average$ to $concept\_similarities$ \tcp*{Store group average}
} 
$Score = \frac{1}{|C|} \sum_{i=1}^{|C|} concept\_similarities[i]$ \tcp*{Calculate the average similarity}
\Return $Score$
\end{algorithm*}

\begin{figure}[h]
\centering
\includegraphics[width=1.0\columnwidth]{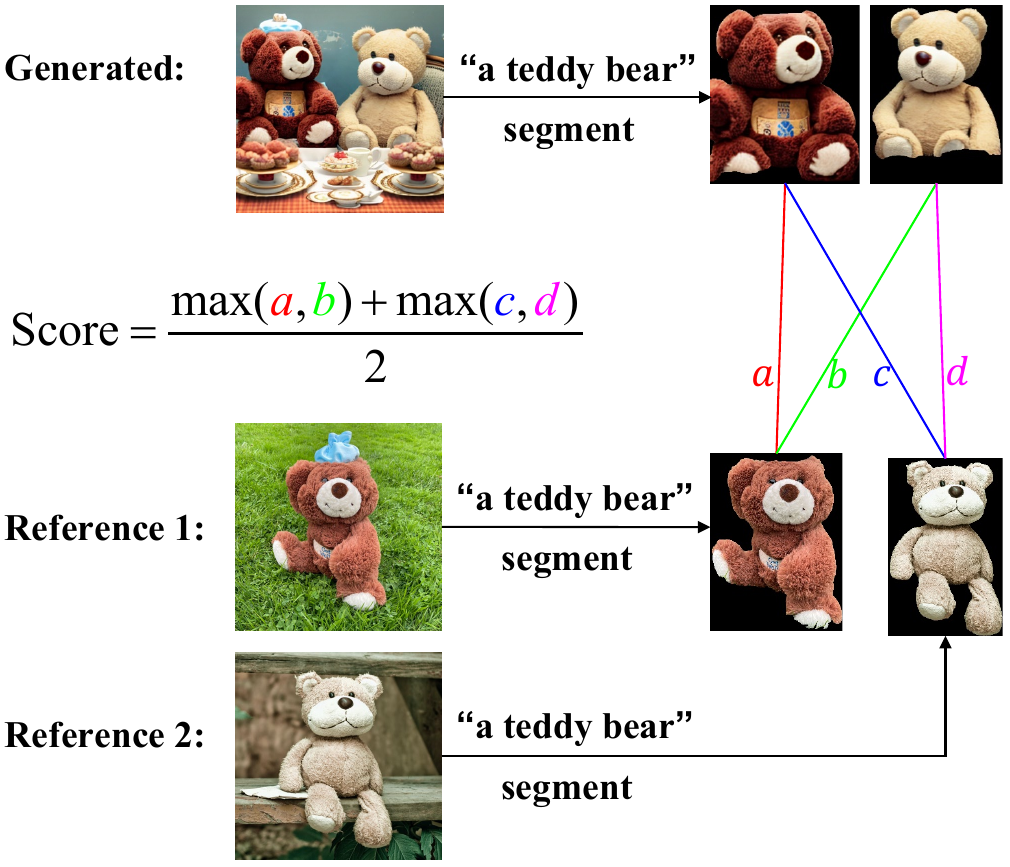} 
\caption{Illustration of our SegSim. a, b, c, and d represent the similarity between two images based on pre-trained scoring models. 
}
\label{fig11}
\end{figure}

\begin{figure*}
\centering
\includegraphics[width=2.0\columnwidth]{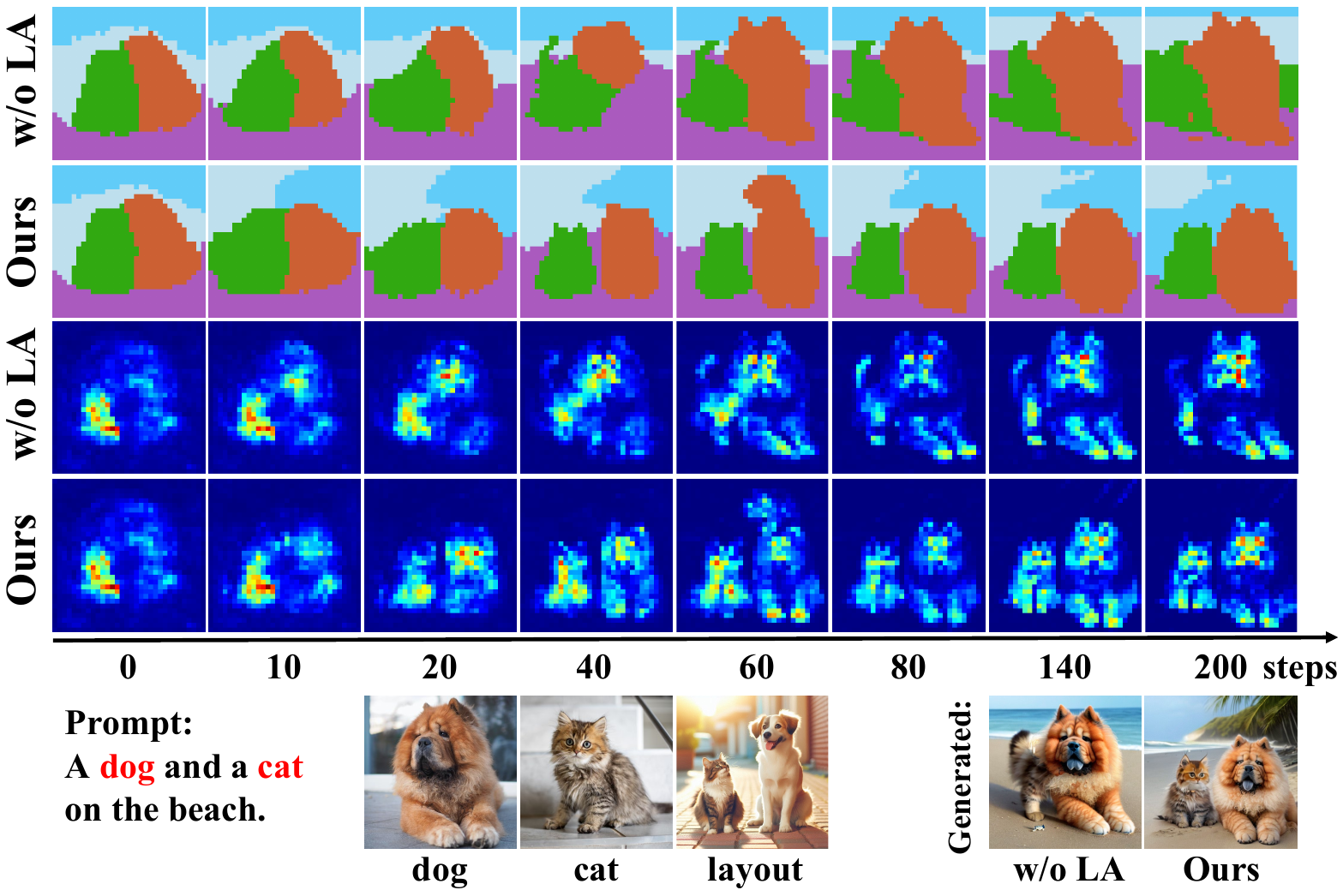} 
\caption{Attention visualization of our method and its variant without layout alignment (LA). Rows 1 and 2 show self-attention visualization. Rows 3 and 4 show cross-attention visualization. The last row displays the input prompt, concepts, layout reference, and generated images.
}
\label{fig12}
\end{figure*}

\begin{figure}[h]
\centering
\includegraphics[width=1.0\columnwidth]{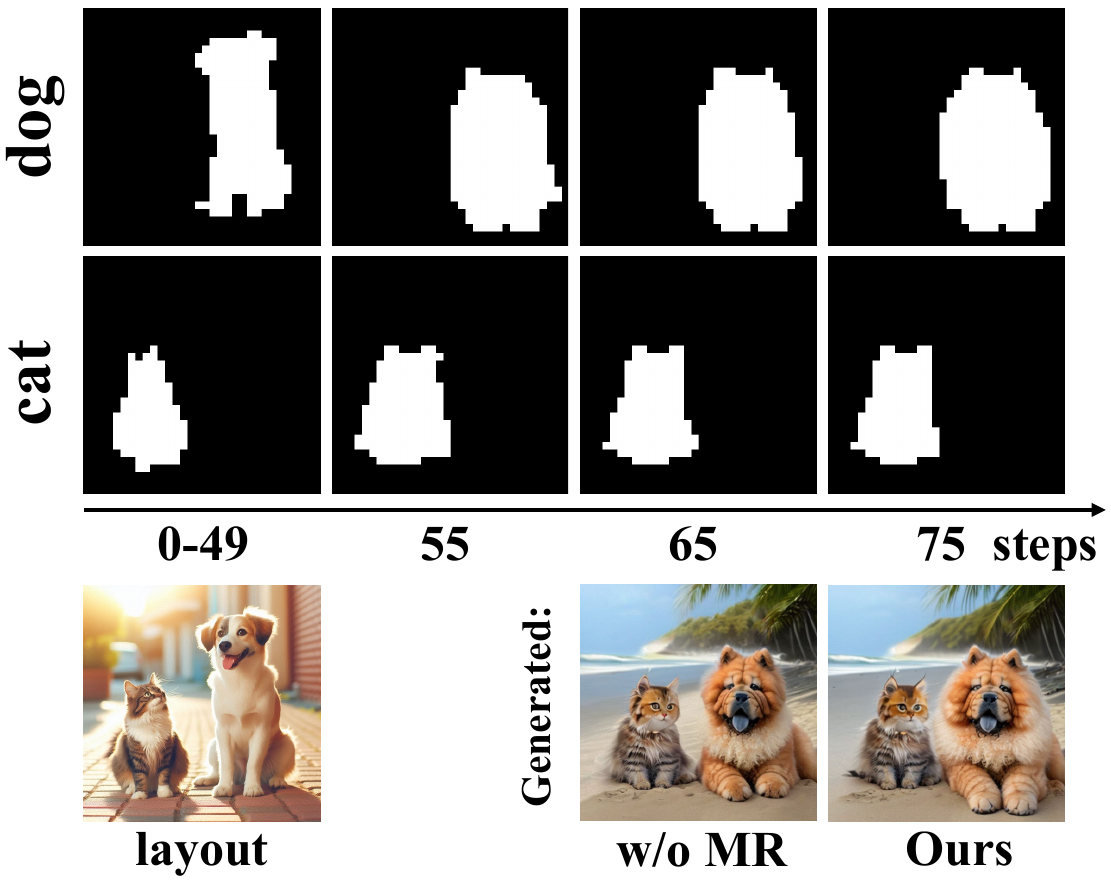} 
\caption{Visualization of masks used for feature fusion. The first row shows masks for the dog, while the second row shows masks for the cat. The last row displays the layout reference, and the images generated by our method and its variant without mask refinement (MR).
}
\label{fig13}
\end{figure}

\begin{figure}
\centering
\includegraphics[width=1.0\columnwidth]{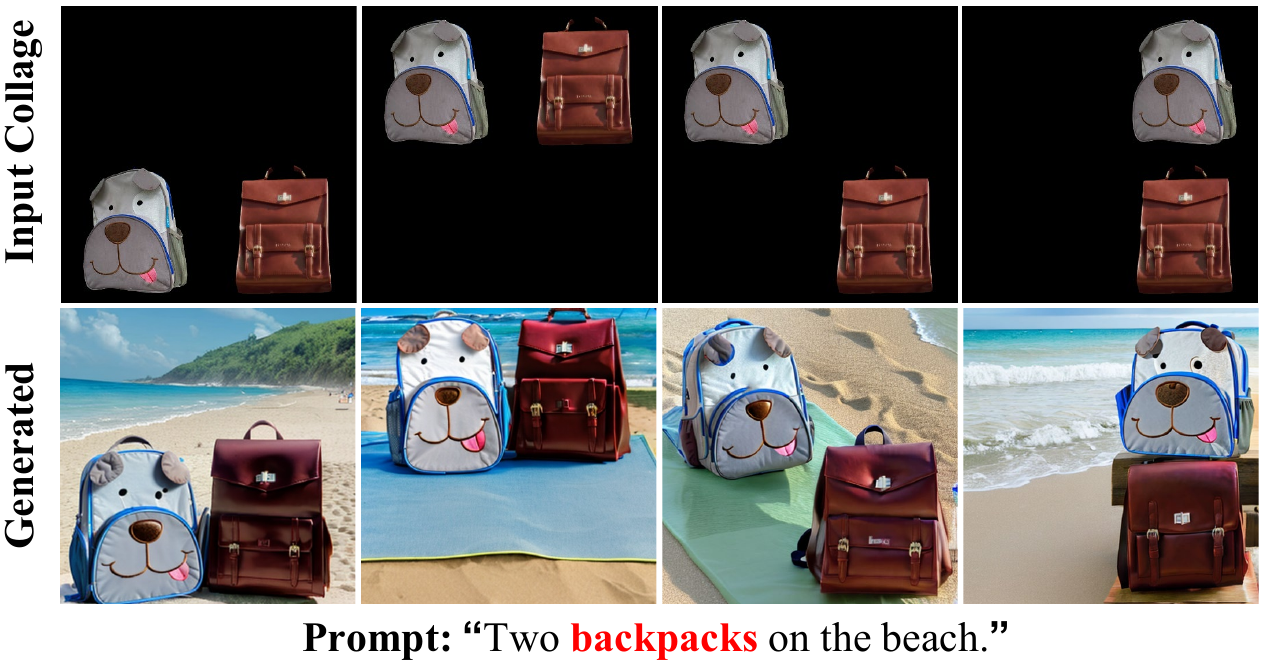} 
\caption{Collage-to-Image Generation. Our method can also utilize a user-created collage as a layout reference and generate images following the given layout. 
}
\label{fig14}
\end{figure}

\begin{figure}
\centering
\includegraphics[width=1.0\columnwidth]{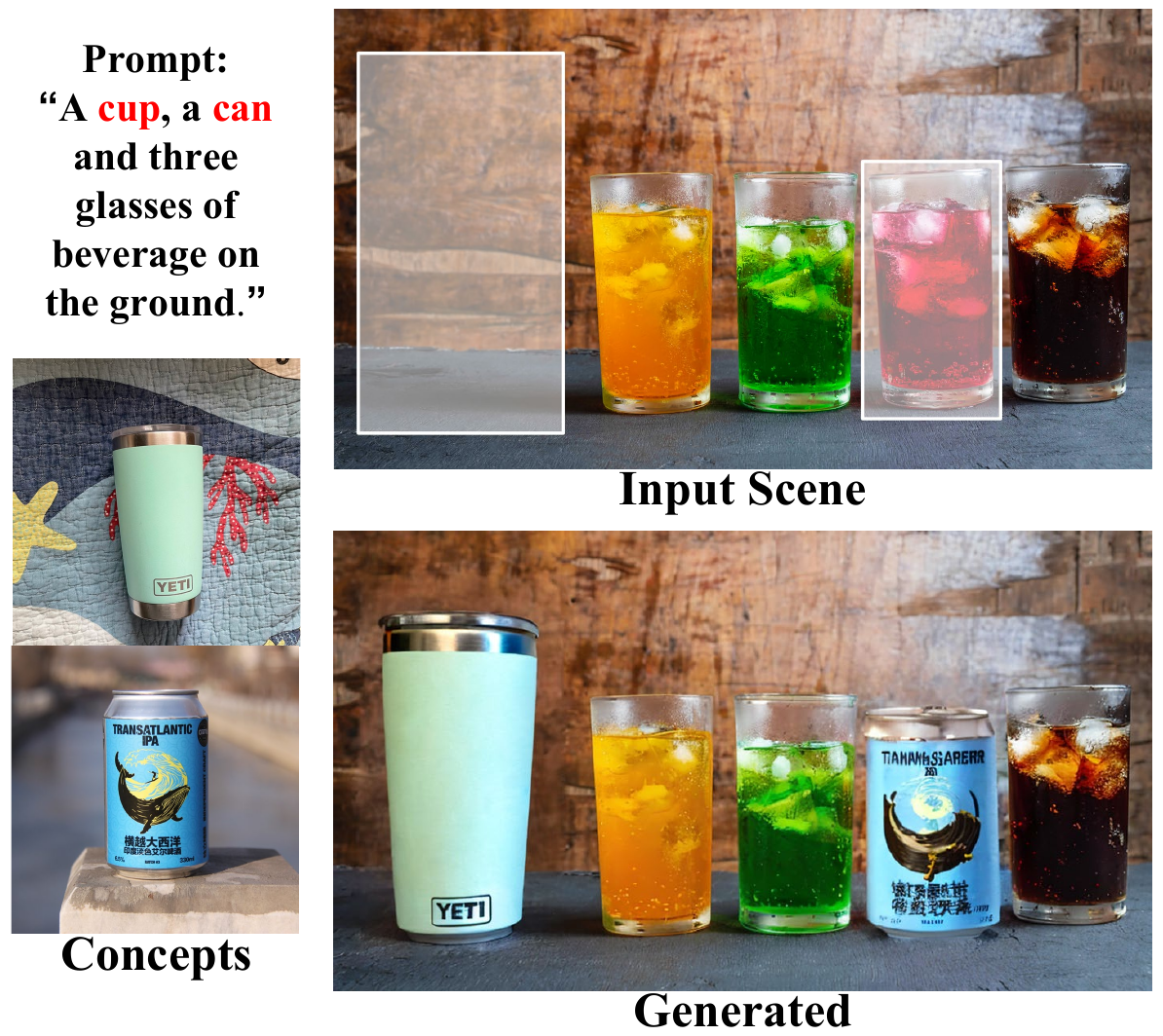} 
\caption{Object Placement. Our method can replace objects in a given scene or add new objects to it.
}
\label{fig15}
\end{figure}

\begin{figure*}
\centering
\includegraphics[width=2.0\columnwidth]{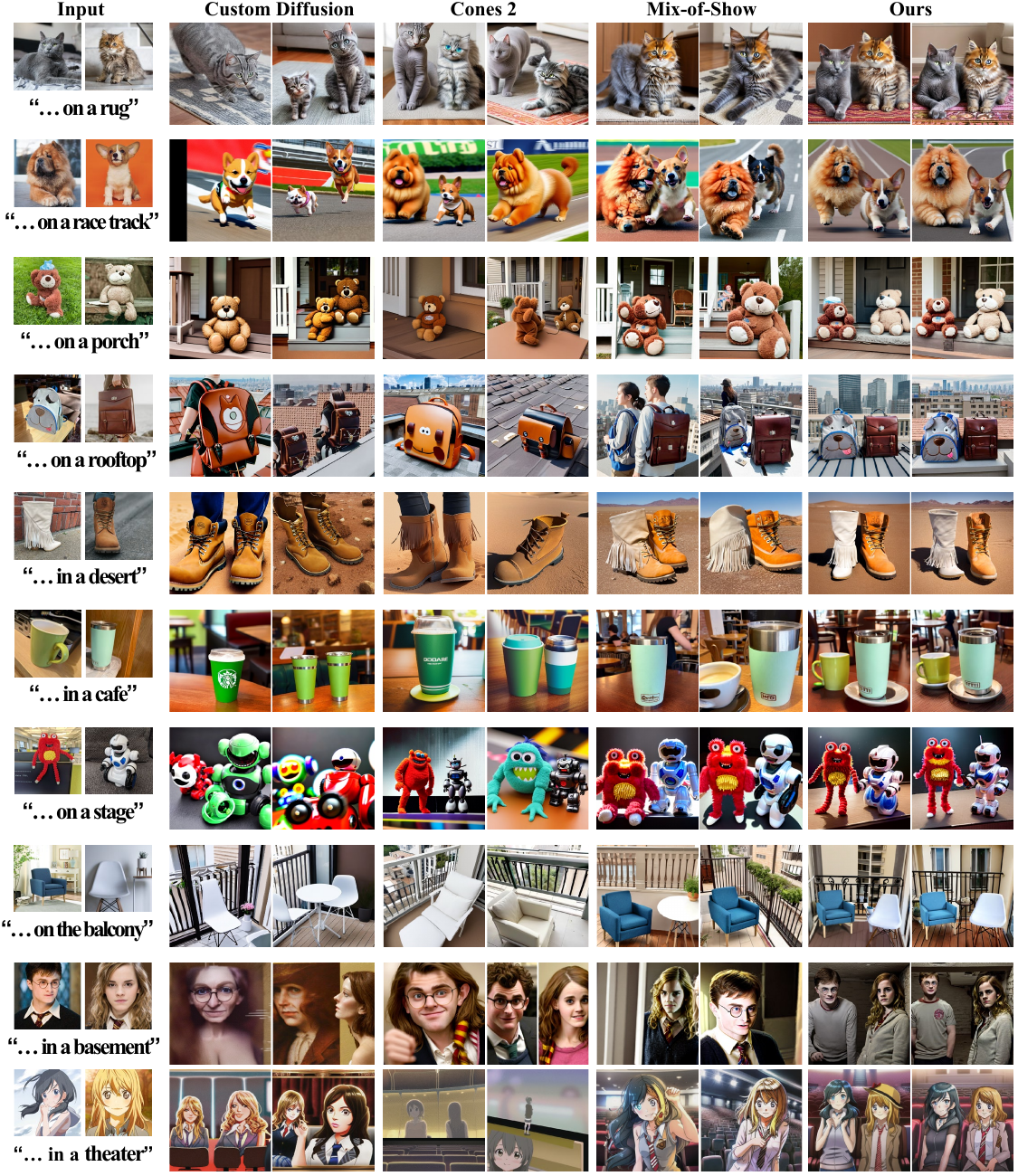} 
\caption{More qualitative comparisons on multi-concept customization. Our method significantly outperforms all baselines in attribute preservation and layout control.
}
\label{fig16}
\end{figure*}

\begin{figure*}
\centering
\includegraphics[width=2.0\columnwidth]{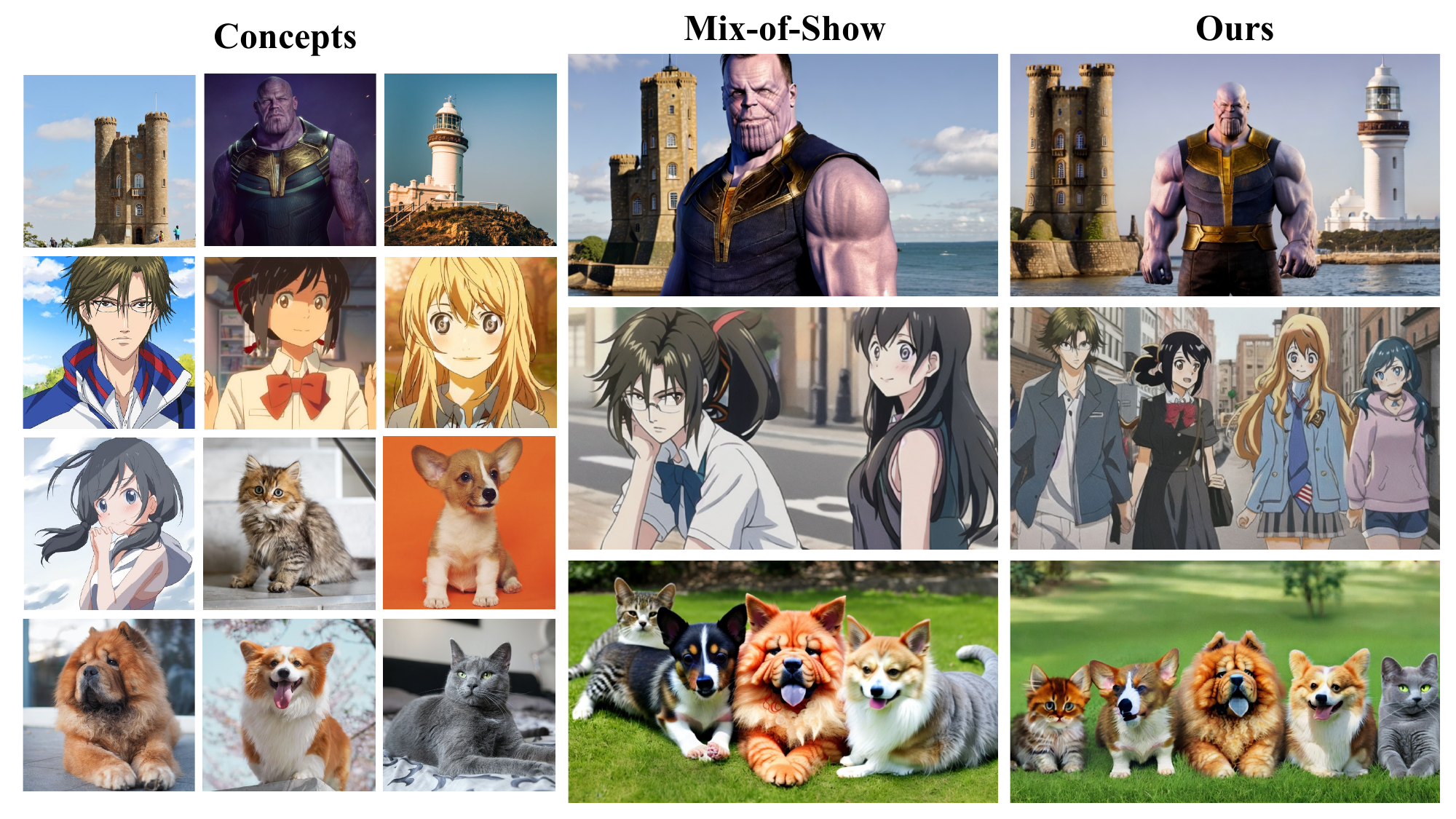} 
\caption{Qualitative comparison on more than two concepts. Our method maintains excellent performance even when handling up to five concepts.
}
\label{fig17}
\end{figure*}

In both qualitative and quantitative comparisons, the original prompts are adapted to fit different methods. For Custom Diffusion, each concept is represented in the “modifier+class” format (e.g., “$<$monster$>$ toy”), resulting in prompts containing two concepts (e.g., “A $<$monster$>$ toy and a $<$robot$>$ toy on a stage.”). For Cones 2, each concept is represented by a two-word phrase (e.g., “monster toy”), leading to prompts with two concepts (e.g., “A monster toy and a robot toy on a stage.”). For Mix-of-Show, each concept is represented by two tokens (e.g., “$<$monster$\_$toy$\_$1$>$ $<$monster$\_$toy$\_$2$>$”), with the original prompt used as the global prompt and two local prompts added (e.g., “A $<$monster$\_$toy$\_$1$>$ $<$monster$\_$toy$\_$2$>$ on a stage” and “A $<$robot$\_$toy$\_$1$>$ $<$robot$\_$toy$\_$2$>$ on a stage”). Our Concept Concept follows the Mix-of-Show representation method but utilizes a base prompt (same as the original prompt) and two prompt variants (e.g., “Two $<$monster$\_$toy$\_$1$>$ $<$monster$\_$toy$\_$2$>$ on a stage” and “Two $<$robot$\_$toy$\_$1$>$ $<$robot$\_$toy$\_$2$>$ on a stage”).

\subsection{A.2\quad Implemental Details}

\subsubsection{Pretrained Models and Sampling. }

We use Stable Diffusion v1.5 as the base model, incorporating pre-trained weights from the community. Following Mix-of-show, we utilize Chilloutmix\footnote{https://civitai.com/models/6424/chilloutmix} for generating real-world concept images and Anything-v4\footnote{https://huggingface.co/xyn-ai/anything-v4.0} for anime concept images. Throughout all experiments detailed in this paper, we apply 200-step DDIM sampling to achieve optimal quality. The classifier-free guidance scale is maintained at 7.5. For quantitative evaluation, we generate 8 images per prompt, with random seeds fixed within the range [0, 7] to ensure reproducibility. All experiments were conducted on an RTX 3090.

\subsubsection{ED-LoRA. }

Following Mix-of-Show\cite{gu2024mix}, we train LoRAs for all attention layers of both the U-Net and text encoder, utilizing Extended Textual Inversion\cite{voynov2023p+} to learn layer-wise embeddings. All training hyperparameters remain consistent with those outlined in the original paper. During inference, the trained LoRA weights are integrated with the pre-trained model weights using a coefficient of 0.7.

\subsubsection{Layout Alignment. }

SDXL\cite{podell2023sdxl} is employed to generate layout reference images according to the base prompt. We perform 1000 steps of DDIM inversion on each layout reference image, recording the self-attention keys at each step as supervision signals. To prevent excessive guidance that may compromise the target concept structure, layout alignment is implemented only from steps 0 to 60.

\subsubsection{Mask Refinement. }

Masks for feature fusion are dynamically adjusted based on the shapes in the self-attention maps. Given $N$ concepts corresponding to $N$ custom models, at time step $t$, clustering is applied to the self-attention maps of each model to extract segmentation maps. Using K-Means clustering with cluster numbers ranging from $N$ to $2N$, all segmentations for each concept are recorded, and the matching degree between each segmentation and the mask at time step $t+1$ is computed. The segmentation with the highest matching degree, defined as the intersection over union of the two masks, is selected as the new mask for the concept in the custom model at time step $t$. Similar operations are performed on the base model to obtain new masks for the $N$ concepts at time step $t$. The new masks from the base model and the corresponding custom models are then combined to form the mask for each concept. To avoid overlap between masks of different concepts, overlapping regions are replaced with the corresponding mask at time step $t+1$. The mask refinement process is detailed in Algorithm 1. Mask refinement is performed every 5 steps from steps 50 to 80, after which the masks for each concept remain unchanged.

\subsection{A.3\quad Segmentation Similarity}

We propose an evaluation metric, Segmentation Similarity (SegSim), to assess the visual consistency between generated images and multiple personalized concepts by calculating image similarity on subject segments. Specifically, Grounded-SAM\cite{ren2024grounded} is used to extract segments of each subject from the generated image using brief prompts (e.g., “a dog” and “a cat”). The same operation is performed on reference images for each target concept. The image similarity between the subject segments of the reference images and that of the generated image is calculated, taking the maximum value as the similarity for that concept. If there are multiple reference images, these results are averaged. Finally, the similarities of all target concepts with the generated image are averaged to obtain the final image alignment score. The overall process of SegSim is illustrated in Figure \ref{fig11} and detailed in Algorithm 2.

\section{B\quad Additional Experiments}

\subsection{B.1\quad Visualizations}

\subsubsection{Visualization of Layout Alignment}

To illustrate our layout alignment process, we visualize the attention probabilities during sampling. For self-attention, we cluster the attention scores of the 6th self-attention layer of the U-Net decoder using K-Means, with different clusters marked in distinct colors. As shown in Figure \ref{fig12}, the shapes of the self-attention regions corresponding to different subjects gradually refine during the denoising process. In early steps, the contours of the self-attention regions are relatively smooth, reflecting the general layout of the image. In later steps, the regions become increasingly complex and irregular, capturing the structural details of the subjects. Consequently, layout alignment is applied only during the first 60 steps to learn the correct layout from the reference image while preserving the structural features of the target concept. By the 60th step, the generated subjects have adopted the shapes of those in the reference image. After ceasing layout alignment, the target subjects gradually revert to their original shapes, while the learned layout is retained.

For cross-attention, we visualize the 5th cross-attention layer of the U-Net decoder by averaging the attention scores of all tokens representing foreground objects. As shown in Figure \ref{fig12}, layout alignment encourages the cross-attention of foreground objects to activate in multiple locations, preventing concept omission or merging. Initially, the attention activation regions are concentrated in the center of the image. During layout alignment, these regions gradually split horizontally into two parts, corresponding to the two target subjects. Layout alignment corrects both self-attention and cross-attention, thus avoiding layout confusion caused by chaotic attention.

\subsubsection{Visualization of Mask Refinement}

We visualize the masks used for feature fusion, as shown in Figure \ref{fig13}. The subject masks are initialized with segmentations extracted from the reference image and remain unchanged during the first 50 steps. Between steps 50 and 80, these masks undergo refinement every 5 steps, after which they remain unchanged. Shortly after refinement begins, the masks’ shapes transition from predefined forms to those of the target concepts. As refinement progresses, the masks make minor adjustments to better fit the contours of the generated subjects. Mask refinement dynamically locates each subject’s area on the attention map, ensuring the visual features of the target concepts are fully injected into the generated image.

\subsection{B.2\quad Applications}

\subsubsection{Collage-to-Image Generation. }

We recommend using real photos or generated images as layout references to achieve reasonable layouts for creating harmonious and natural images. However, existing image layouts may not always align with user preferences. To address the need for uncommon or complex layouts, our method allows users to create a collage as a reference image, precisely describing their desired layout. This collage can be easily created with the assistance of powerful segmentation models (e.g., SAM\cite{kirillov2023segment}). As shown in Figure \ref{fig14}, our method generates harmonious images based on the layouts of the collages, preserving the visual details of custom concepts even if the layouts are unconventional.

\subsubsection{Object Placement. }

Our method can be combined with inpainting techniques\footnote{https://huggingface.co/docs/diffusers/using-diffusers/inpaint} to replace objects in a given image or add new ones. At each denoising step, DDIM inversion converts the image to be edited into a latent space representation. Spatial fusion is then performed between the inverted latent vectors and those generated by our Concept Conductor, based on a user-defined mask. For object replacement, the image to be edited serves as the layout reference. For object addition, the segmentation of the target concept is pasted onto the original image as the layout reference. As illustrated in Figure \ref{fig15}, our method seamlessly places multiple custom concepts in the target locations of a given scene.

\subsection{B.3\quad User Study}

\begin{table}[h]
\begin{tabular}{lcc}
\hline Method & Text-alignment & Image-alignment \\
\hline Custom Diffusion & 1.7600 & 1.5725 \\
Cones 2 & 2.3899 & 1.3474 \\
Mix-of-Show & 2.1950 & 2.3200 \\
Ours & $\mathbf{4.7400}$ & $\mathbf{4.3750}$ \\
\hline
\end{tabular}
\caption{User Preference Study. Our method is the most favored by users, receiving the highest ratings for both text alignment and image alignment.}
\label{table3}
\end{table}

We conduct a user study to further evaluate our method. We assess human preferences for generated images in two aspects: (1) \textbf{Text Alignment}. Participants are shown images generated by different methods along with the corresponding text prompts. They rate how well the generated images match the text descriptions on a scale from 1 (not at all) to 5 (completely), representing text alignment. (2) \textbf{Image Alignment}. Participants are shown the generated images and reference images for multiple target concepts. They rate the similarity between the generated images and each target concept on a scale from 1 (not at all) to 5 (very similar). The average similarity scores for different concepts are used as image alignment. If a concept is absent in the generated image, participants are asked to give the lowest score. We collected feedback from 20 users, each evaluating 40 generated images. As shown in Table \ref{table3}, our method significantly outperforms the baselines in both text and image alignment, consistent with the results of automatic evaluations.

\subsection{B.4\quad More Qualitative Comparisons}

Figure \ref{fig16} presents additional qualitative results comparing our method with the baselines. Our method ensures correct attributes and layouts across various scenarios, while the baselines suffer from severe attribute leakage and layout confusion. Furthermore, we compare the performance of Mix-of-Show\cite{gu2024mix} and our method when handling more than two concepts, as shown in Figure \ref{fig17}. As the number of concepts increases, Mix-of-Show encounters significant issues with missing or mixed subjects. In contrast, our method accurately generates all concepts without confusion, demonstrating its effectiveness in attribute preservation and layout control.

\section{C\quad Limitations}

Firstly, our method encounters quality issues when generating small subjects. For instance, generated small faces may become distorted and deformed. This issue, also observed in Mix-of-Show\cite{gu2024mix}, is primarily due to the VAE losing visual details of the target subjects when compressing the image information. Replacing the base model, increasing image resolution, or changing the layout may help address this problem.

Secondly, our method incurs considerable computational overhead. To avoid high memory usage from parallel sampling of multiple custom models, we alternately load different concepts’ ED-LoRA weights at each timestep, which reduces inference efficiency. Additionally, performing backpropagation during sampling to update latent representations further increases latency.

\section{D\quad Social Impacts}

Our Concept Conductor demonstrates significant innovation and potential in text-to-image generation, particularly in multi-concept customization. Our method generates images with correct layouts that include all target concepts while preserving each concept’s original characteristics and visual features, avoiding layout confusion and attribute leakage. This technology can provide users with more efficient creative tools, inspiring artistic exploration and innovation, and potentially impacting fields such as advertising, entertainment, and education. However, this powerful image generation capability could also be misused for unethical activities, including image forgery, digital impersonation, and privacy invasion. Therefore, it is recommended to incorporate appropriate ethical reviews and safeguards to prevent potential misuse and harm to the public. Future research should continue to address these ethical and security issues to ensure the technology’s proper and responsible use.

\end{document}